\documentclass[11pt]{article}

\usepackage[final]{acl}

\usepackage{float}
\usepackage[table]{xcolor}
\usepackage{caption}
\usepackage{subcaption}
\usepackage{times}
\usepackage{latexsym}
\usepackage{float}
\usepackage{amsmath}
\usepackage{amssymb}
\usepackage{mathrsfs} 
\usepackage[most,skins,theorems]{tcolorbox}  
\usepackage{xcolor}      
\usepackage{multirow}
\usepackage{pifont}
\usepackage{booktabs}
\tcbuselibrary{breakable,skins}
\usepackage[T1]{fontenc}
\usepackage{enumitem}
\usepackage{afterpage}
\usepackage[utf8]{inputenc}

\usepackage{microtype}
\usepackage{makecell}
\usepackage{stfloats}
\usepackage{inconsolata}
\usepackage{caption}
\usepackage{graphicx}
\usepackage{titlesec}
\usepackage{multirow}
\usepackage[table,xcdraw]{xcolor}
\usepackage[normalem]{ulem}
\useunder{\uline}{\ul}{}

\usepackage[utf8]{inputenc}

\usepackage{microtype}

\usepackage{inconsolata}

\usepackage{graphicx}
\usepackage{hyperref}
\usepackage{url}
\usepackage{enumitem}
\usepackage{graphicx}
\usepackage{wrapfig}
\usepackage{multirow}
\usepackage{adjustbox} 
\usepackage{graphicx}  
\usepackage{bm}
\usepackage{xcolor}
\usepackage{colortbl}
\usepackage{booktabs}
\usepackage[dvipsnames]{xcolor}      
\usepackage{amsmath}
\usepackage{amssymb}
\usepackage[ruled,vlined]{algorithm2e}
\usepackage{amsfonts,amssymb}
\usepackage{subcaption}
\usepackage[utf8]{inputenc}
\usepackage{microtype}
\usepackage{float}

\tcbset{
  promptboxstyle/.style={
    width=\linewidth,
    top=8pt,
    bottom=4pt,
    colback=white!10!white,
    colframe=black,
    colbacktitle=black,
    enhanced,
    breakable,
    arc=3mm,
    frame hidden,
    borderline={1pt}{0pt}{black},
    skin first is subskin of={emptyfirst}{%
      frame code={\draw[black,line width=1pt,rounded corners=3mm] 
        (frame.south west) rectangle (frame.north east);}
    },
    skin middle is subskin of={emptymiddle}{%
      frame code={\draw[black,line width=1pt] 
        (frame.south west) rectangle (frame.north east);}
    },
    skin last is subskin of={emptylast}{%
      frame code={\draw[black,line width=1pt,rounded corners=3mm] 
        (frame.south west) rectangle (frame.north east);}
    },
    center,
    attach boxed title to top left={yshift=-0.1in,xshift=0.15in},
    boxed title style={boxrule=0pt,colframe=white,},
    fonttitle=\small\bfseries,
    before upper={\fontsize{10}{12}\selectfont},
  }
}
\newtcolorbox{Promptbox}[2][]{promptboxstyle,title=#2,#1}

\title{Scaling LLM Knowledge Boundaries via Distribution-Optimized Synthesis}

\author{
    Songze Li\textsuperscript{1,3},
    Yarong Lan\textsuperscript{1,3},
    Zhongpu Bo\textsuperscript{2}, 
    Zhaoyang Wang\textsuperscript{2}, \\
    \textbf{Zhiqiang Liu}\textsuperscript{1},
    \textbf{Yuan Yuan}\textsuperscript{1},
    \textbf{Chengtao Gan}\textsuperscript{1},
    \textbf{Menghao Qian}\textsuperscript{1},  
    \textbf{Enpei Niu}\textsuperscript{1}, \\
    \textbf{Xiaoke Guo}\textsuperscript{1}, 
    \textbf{Yuanxiang Liu}\textsuperscript{1}, 
    \textbf{Zhaoyan Gong}\textsuperscript{1,3}, 
    \textbf{Xiangjin Hu}\textsuperscript{1,3},
    \textbf{Liangyurui Liu}\textsuperscript{1},
    \textbf{Jingdian Lu}\textsuperscript{1,3}, \\
    \textbf{Lei Liang}\textsuperscript{2},
    \textbf{Jun Zhou}\textsuperscript{2},
    \textbf{Huajun Chen}\textsuperscript{1},
    \textbf{Wen Zhang\textsuperscript{1,3}\thanks{~~Corresponding authors.}}\\
    \textsuperscript{1}Zhejiang University,
    \textsuperscript{2}Ant Group,
    \textsuperscript{3}ZJU-Ant Group Joint Lab of Knowledge Graph \\
    \texttt{
    \{li.songze,zhang.wen\}@zju.edu.cn
    }
}

\begin{document}
\maketitle
\begin{abstract}
Knowledge injection via synthetic data is crucial for enhancing Large Language Models (LLMs). However, current synthesis methods simply stop at preset token counts or fixed data ratios, lacking awareness of knowledge distribution. This results in some domains being sparse while others are redundant, limiting LLM knowledge boundaries. We revisit knowledge injection from a distribution perspective and hypothesize that an optimal knowledge distribution exists to maximize knowledge boundary expansion. We propose \textbf{KDoS} (\textbf{K}nowledge \textbf{D}istribution-\textbf{o}ptimized \textbf{S}ynthesis), a framework that introduces knowledge density to drive synthesis through a three-stage feedback mechanism, shifting from blind generation to distribution-optimized synthesis. We construct Wikipedia-based synthetic data with varying knowledge distributions and conduct experiments on models from 0.6B to 16B (Qwen, Ling, LLaMA) and data scales from 1B to 5B tokens. Our key findings are: (1) an optimal knowledge distribution consistently maximizes boundary expansion; (2) this distribution is stable across backbones and scales; (3) KDoS outperforms baselines across six knowledge benchmarks. Our work offers a new perspective and practical framework for synthetic data-driven knowledge injection.
\end{abstract}

\section{Introduction}

The knowledge boundary of an LLM defines the scope of knowledge it can reliably handle \citep{li-etal-2025-knowledge-boundary}, serving as a core dimension for assessing model capabilities \citep{zhao-etal-2025-know,yin2023largelanguagemodelsknow}. However, even state-of-the-art LLMs exhibit knowledge boundary limitations—particularly on long-tail knowledge \citep{10.5555/3618408.3619049} such as infrequent Wikipedia facts—where low-frequency but factually grounded questions cannot be reliably answered \citep{sun-etal-2024-head,mallen-etal-2023-trust}. Scaling up LLM knowledge boundaries through efficient knowledge injection has thus become an important research challenge \citep{10.5555/3524938.3525306, allenzhu2024physicslanguagemodels33}. 

\begin{figure*}[ht]
    \centering
    \includegraphics[width=1.0\textwidth, keepaspectratio]{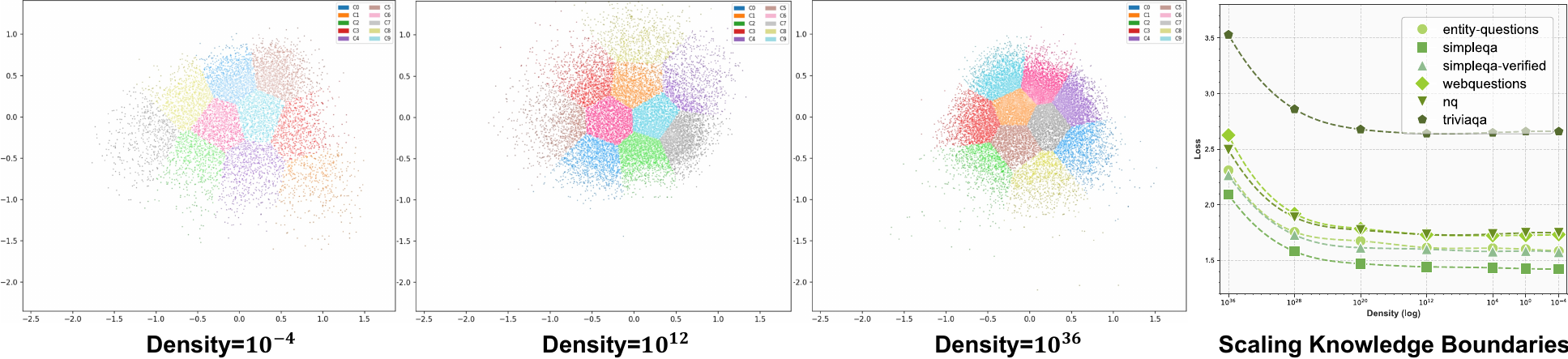} 
    \caption{Scaling LLM Knowledge Boundaries with Different Distributions. We present scatter plots of three different knowledge densities ($10^{-4}$, $10^{12}$, and $10^{36}$) (computed via Eq.~\ref{eq:density}), along with eval loss scaling curves on six knowledge benchmarks. The loss gap between $10^{-4}$ and $10^{36}$ reaches up to 24.6\%.} 
    \label{fig:motivation}
\end{figure*}

Synthetic data offers a flexible, scalable, and cost-effective way to target specific knowledge domains \citep{ke2023continualpretraininglanguagemodels}, alleviating long-tail coverage gaps in real data, and has been widely adopted in continual pre-training, instruction tuning, and other knowledge injection settings \citep{10.5555/3666122.3666237,yang2024syntheticcontinuedpretraining}. However, existing methods share a fundamental limitation: they simply stop at preset token counts or fixed data ratios, with no awareness or control over knowledge distribution \citep{azerbayev2024llemmaopenlanguagemodel,10.1145/3701716.3715245}. This constitutes a blind-synthesis paradigm: methods neither perceive the current knowledge distribution nor understand which distribution optimizes injection efficiency. As a result, some knowledge points are redundantly repeated while others remain critically sparse \citep{havrilla2024surveyingeffectsqualitydiversity}, directly constraining the model's knowledge boundary \citep{10.5555/3666122.3667604}. As illustrated in Fig. \ref{fig:motivation}, training the same model with and equal number of tokens but different knowledge distributions leads to a loss difference of up to 24.6\% on knowledge QA benchmarks, confirming that knowledge distribution is a key factor in injection effectiveness \citep{10.5555/3737916.3738886}.

To address this limitation, we revisit LLM knowledge injection from a distribution perspective and propose a core hypothesis: \textit{there exists at least one optimal knowledge distribution that maximizes knowledge boundary expansion}. Based on this, we propose \textbf{KDoS} (\textbf{K}nowledge \textbf{D}istribution-\textbf{o}ptimized \textbf{S}ynthesis), which shifts synthetic data generation from blind synthesis to targeted distribution shaping. KDoS introduces knowledge density as a controllable proxy for knowledge distribution, operating through three iterative stages: (1) extracting and extending knowledge points from seed knowledge, organizing semantically related samples into knowledge groups, and generating new questions from knowledge point combinations within each group; (2) quality filtering of candidate questions; (3) rejection sampling over candidates based on a target knowledge distribution. KDoS continuously monitors the knowledge distribution of the data pool and dynamically adjusts the acceptance strategy, iteratively driving the distribution toward the preset target.

Our main contributions are as follows:

\begin{itemize}[leftmargin=*]
    \item \textbf{Problem Perspective.} We reframe knowledge injection as a knowledge distribution control problem, identifying the lack of distribution awareness as the fundamental limitation of existing methods, and offering a new research perspective for the field.
    \item \textbf{Methodological Innovation.} We propose KDoS, which introduces knowledge density as a controllable variable for knowledge distribution, and employs a dynamic feedback mechanism to precisely shape the distribution of synthetic data, enabling continuous scaling of LLM knowledge boundaries.
    \item \textbf{Experimental Insights and Validation.} Through systematic experiments across LLMs from 0.6B to 16B and varying data scales, we confirm our hypothesis and find that the optimal knowledge density consistently exists across different backbones and data scales, manifesting as $10^{-4}$\textasciitilde$10^{4}$ on our data and revealing a general principle of knowledge injection. Extensive experiments further validate that KDoS consistently outperforms existing baselines across six knowledge benchmarks.
\end{itemize}

\section{Related Work}

\paragraph{LLM Knowledge Injection.} Scaling up the parametric knowledge boundaries of LLMs is a core challenge for improving their fundamental capabilities. Knowledge injection approaches include continual pre-training, supervised fine-tuning (SFT), and retrieval-augmented generation. ADEPT \citep{zhang2025adept} and LLaMA-Pro \citep{wu2024llama} extend model architecture for continual pre-training, but \citep{lv2025inject} identify a ``memory collapse'' threshold in knowledge injection, revealing inherent limitations of pre-training approaches. \citep{ovadia2024fine} show that retrieval-augmented methods can outperform fine-tuning in certain scenarios without training, while knowledge editing methods such as ROME \citep{meng2022locating} and MEMIT \citep{meng2022mass} face scalability bottlenecks under large-scale updates.
As for LLM knowledge evaluation, knowledge-intensive benchmarks such as TriviaQA \citep{joshi2017triviaqa}, Natural Questions \citep{kwiatkowski2019natural}, and WebQuestions \citep{talmor2018webknowledgebaseansweringcomplex} primarily assess common factual knowledge. SimpleQA \citep{wei2024measuring}, SimpleQA-Verified \citep{haas2025simpleqa}, and Entity Questions \citep{sciavolino2021simple} target long-tail knowledge, where even state-of-the-art models perform poorly. \citep{10.5555/3618408.3619049} further show that QA performance strongly correlates with document frequency in pre-training data, confirming that LLMs systematically struggle with long-tail knowledge. These studies highlight that existing knowledge injection methods leave significant knowledge gaps in long-tail settings such as Wikipedia, and efficiently expanding LLM knowledge boundaries remains an open problem.

\paragraph{Data Synthesis.} For general-purpose data synthesis, Self-Instruct \citep{wang2023self}, Evol-Instruct \citep{xu2023wizardlm}, and Magpie \citep{xu2025magpie} establish foundational paradigms for instruction synthesis, but all terminate synthesis at a human-specified token count without considering data distribution. For quality and distribution control, STaR \citep{zelikman2022starbootstrappingreasoningreasoning}, RFT \citep{yuan2023scalingrelationshiplearningmathematical}, DART-Math \citep{tong2024dartmathdifficultyawarerejectiontuning}, DEITA \citep{liu2024makesgooddataalignment}, and TreeSynth \citep{wang2025treesynthsynthesizingdiversedata} improve synthesis from the perspectives of reasoning, sampling, and diversity, yet still lack explicit modeling of knowledge distribution. For knowledge-aware synthesis, GraphGen \citep{chen2025graphgenenhancingsupervisedfinetuning} and CodecLM \citep{wang-etal-2024-codeclm} incorporate knowledge graphs and metadata to guide generation, but neither actively controls knowledge distribution during synthesis. \citep{Qin2025ScalingLO} identify performance bottlenecks in synthetic data but offer no mechanism to dynamically adjust knowledge distribution.
In summary, existing data synthesis methods remain fundamentally blind to knowledge distribution. We therefore propose KDoS, which scales LLM knowledge boundaries by optimizing the knowledge distribution of synthetic data.

\section{Methods}

\subsection{Preliminary}

\label{Knowledge Density Definition}
\paragraph{Knowledge Density Definition.}
We denote the data pool as $\mathcal{S} = (T, \rho)$, where $T$ and $\rho$ represent the token count and knowledge density of $\mathcal{S}$, respectively. Following \citep{chen-etal-2025-revisiting}, we define $\rho$ as:
\vspace{-10pt}
\begin{equation}
\label{eq:density}
\scalebox{0.9}{$\displaystyle
\rho = \frac{T}{V} = \frac{T \cdot \Gamma(n/2 + 1)}{\pi^{n/2} \cdot r^{n}},
$}
\end{equation}
where $V$ is the volume of the $n$-dimensional hypersphere formed by $\mathcal{S}$ in semantic space, and $r$ is the average radius, i.e., the mean distance from all samples to the centroid. A higher $\rho$ indicates that more knowledge is concentrated in a smaller semantic region, while a lower $\rho$ indicates sparser coverage over a broader semantic space.

\paragraph{Problem Definition.}
Given a seed question pool $\mathcal{S}^{\text{seed}}$, a synthesis method $\mathcal{A}$ produces synthetic data $\mathcal{S}^{\text{syn}}$ (i.e., \(\left. \mathcal{A}\left( \mathcal{S}^{seed},\mathcal{D}^{target} \right)\rightarrow\mathcal{S}^{syn} \right.\)), which is used to train LLM $\mathcal{M}$ and evaluated on test set. The goal is to maximize the test accuracy $\text{Acc}(\mathcal{M}, \mathcal{S}^{\text{syn}})$ to achieve optimal knowledge injection and expand the knowledge boundary of $\mathcal{M}$. Formally:
\vspace{-5pt}
\begin{equation}
\scalebox{0.9}{$\displaystyle
\underset{\mathcal{D}}{\left. \mathcal{D}^{*} \right.\sim{argmax}}{{\mathbb{E}}_{P(\mathcal{S}^{syn})\sim\mathcal{D}}\left\lbrack {Acc\left( \mathcal{M},\mathcal{S} \right.}^{syn} \right)\rbrack},
$}
\end{equation}
where $\mathcal{D}^{*}$ is the optimal knowledge distribution.
Given a target token count $T^{\text{target}}$ and target density $\rho^{\text{target}}$, KDoS iteratively drives the data pool to converge to the target distribution $\mathcal{D}^{\text{target}}$ over $k$ iterations:
\vspace{-5pt}
\begin{equation}
\scalebox{0.9}{$\displaystyle
\begin{aligned}
\mathcal{D}_{t+1} &\leftarrow \text{Update}\left[\mathcal{D}_{t}, \mathcal{A}_{t}\left(\mathcal{S}_{t}, \mathcal{D}^{\text{target}}\right)\right], \\
&\phantom{{}\leftarrow{}} \text{s.t.} \quad \lim_{t \rightarrow k} \mathcal{D}_{t} = \mathcal{D}^{\text{target}}
\end{aligned}
$}
\end{equation}

\subsection{Overview of Data Synthesis and Verification}

\begin{figure*}[ht]
    \centering
    \includegraphics[width=1.0\textwidth, keepaspectratio]{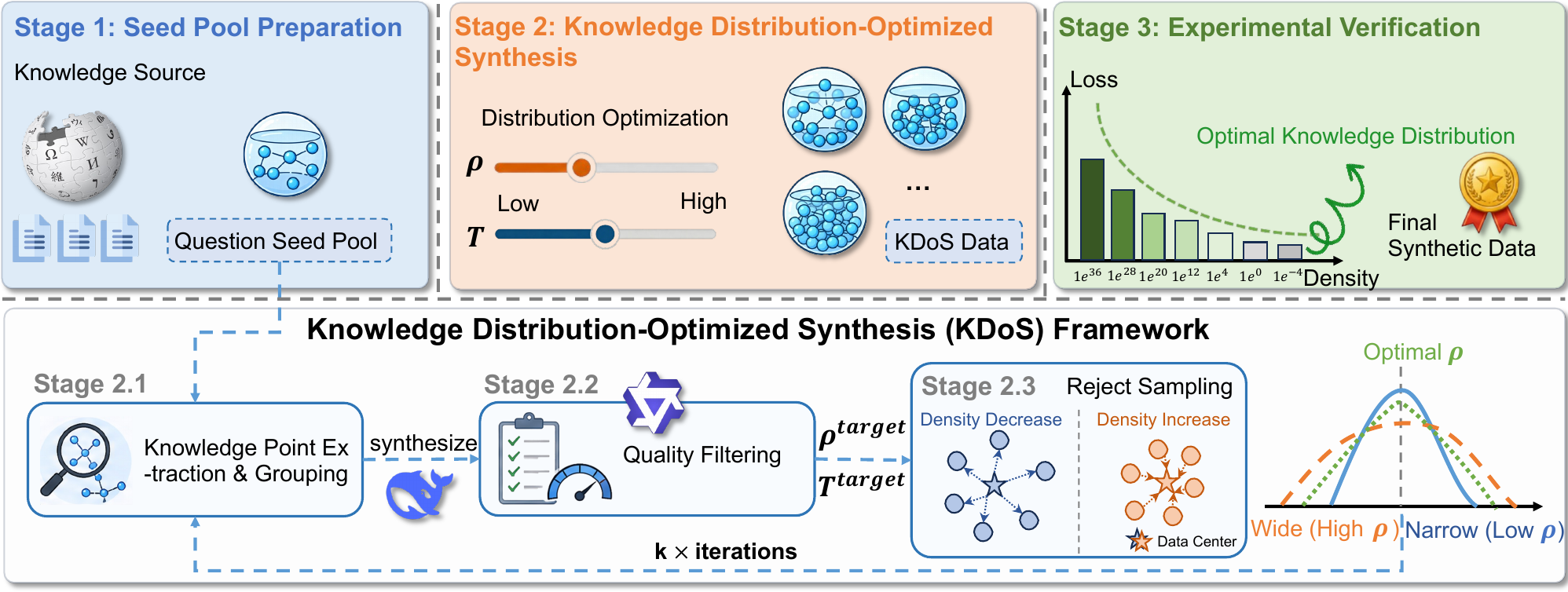} 
    \caption{\textbf{Top:} Overview of our data synthesis and verification pipeline. \textbf{Bottom:} Overview of the \textbf{KDoS} (\textbf{K}nowledge \textbf{D}istribution-\textbf{o}ptimized \textbf{S}ynthesis) framework.} 
    \label{fig:methods}
\end{figure*}

Our work focuses on scaling LLM knowledge boundaries and consists of three stages: 
\textbf{Stage 1: Seed Pool Preparation}, which collects and processes raw data; 
\textbf{Stage 2: Knowledge Distribution-Optimized Synthesis (KDoS)}, our proposed framework for distribution-optimized synthesis; 
\textbf{Stage 3: Experimental Verification}, which evaluates the performance of different knowledge distributions $\mathcal{D}$ on LLM knowledge injection.

\paragraph{Seed Pool Preparation.}
To support large-scale data synthesis, we collect raw documents from Wikipedia. After data cleaning, we summarize knowledge points from each document and synthesize seed questions whose answers are directly grounded in the source documents. This step yields approximately 14M seed QA pairs. Details are provided in Appendix \ref{Seed Pool Preparation}.

\paragraph{Knowledge Distribution-Optimized Synthesis.}
Based on the seed question pool from Stage 1, KDoS controls the synthesis process according to preset $T^{\text{target}}$ and $\rho^{\text{target}}$, driving the final synthetic data $\mathcal{S}^{\text{syn}} = (T^{\text{target}}, \rho^{\text{target}})$ to conform to the target distribution $P(\mathcal{S}^{\text{syn}}) \sim \mathcal{D}^{\text{target}}$. This is described in detail in Section \ref{KDoS Framework}.

\paragraph{Experimental Verification.}
We evaluate synthetic data of varying knowledge distributions from Stage 2 on six knowledge benchmarks, examining how LLM knowledge boundaries scale with knowledge distribution, identifying the optimal knowledge density range, and exploring general principles of knowledge injection across multiple model sizes, data scales, and backbones.

\subsection{KDoS Framework}
\label{KDoS Framework}
KDoS operates through three iterative stages. First, KDoS extracts a knowledge point list and knowledge logic chain for each question, maps them to a semantic space to form knowledge groups based on semantic proximity, and synthesizes new candidate questions via knowledge combination. Second, KDoS applies quality filtering to remove low-quality samples. Third, KDoS uses rejection sampling to preferentially select samples that drive the data pool toward the target distribution, iterating until convergence.

\subsubsection{Knowledge Point Extraction \& Grouping}
For each question in the seed pool $\mathcal{S}^{\text{seed}}$, we use DeepSeek V3.2 \citep{DBLP:journals/corr/abs-2512-02556} to extract a knowledge point list and a knowledge logic chain. The knowledge point list captures the relevant knowledge required to answer the question, while the knowledge logic chain provides an explicit representation of the logical relationships among knowledge points. Based on the knowledge point lists, we apply $n$-gram ($n=5$) deduplication on knowledge point lists, retaining only one question among those with similar knowledge points. We also perform overlap detection between the current data pool and the test set to exclude samples that may cause test set leakage.
We then map each knowledge point list to a semantic space using the embedding model \texttt{sentence-transformers/all-MiniLM-L6-v2} \citep{reimers2019sentencebertsentenceembeddingsusing}, which enables grouping semantically similar samples into knowledge groups. Each knowledge group consists of a sample and its two nearest neighbors in semantic space, forming a group of 3 samples. For each group, we use DeepSeek V3.2 to semantically extend the knowledge points within the group (i.e., extending new related knowledge points from the model's parametric knowledge), and then combine knowledge points across the group to synthesize 5 new candidate questions. This promotes greater knowledge diversity, breaks the knowledge boundary of individual seed questions, and broadens knowledge coverage. Examples of knowledge point lists, logic chains, and knowledge groups are provided in Appendix \ref{Knowledge Point Extraction and Grouping}; prompts for knowledge extraction, semantic extension, and question synthesis are in Appendix \ref{Knowledge Points Extraction Prompt}, \ref{Synthetic Prompt}.

\subsubsection{Quality Filtering}

Synthesized candidate questions suffer from issues like ambiguous intent, meaningless content, hallucinated answers, incorrect knowledge points, etc. To address this, we use DeepSeek V3.2 and Qwen3.5-397B-A17B \citep{DBLP:journals/corr/abs-2604-15804} as LLM judges in a two-step evaluation.

\paragraph{Preliminary Check.} Each sample undergoes three binary tests: \textbf{Answer Independence} verifies that the answer is not directly inferable from the question itself; \textbf{Answer Verifiability} requires objective and verifiable answers; \textbf{Answer Correctness} filters out factual or common-sense errors. Samples failing any criterion are discarded immediately.

\paragraph{Scoring.} Samples passing the preliminary check are scored across five dimensions (total: 12 points): \textbf{Educational Significance (0\textasciitilde4)} penalizes trivial or content-free questions; \textbf{Specificity and Concreteness (0\textasciitilde2)} encourages instance-level questions over abstract ones; \textbf{Internal Question Logic (0\textasciitilde2)} checks coherence of the question itself; \textbf{Question-Answer Logic (0\textasciitilde2)} ensures the answer logically follows from the question; \textbf{Knowledge-Point Relevance \& Logic Completeness (0\textasciitilde2)} verifies that the associated knowledge points are relevant and form a complete reasoning chain. Samples scoring zero on any dimension are excluded, and only those with an average score $\geq 8$ from both judges are retained. The LLM judge prompt is in Appendix \ref{Evaluation Prompt}.

\subsubsection{Reject Sampling}
\label{Reject Sampling}
In this stage, KDoS applies rejection sampling to iteratively drive the current distribution $\mathcal{D}_{t}$ toward $\mathcal{D}^{\text{target}}$. The process consists of two phases:
\textbf{Cold-start phase} ($T < T^{\text{target}}$): The goal is to accumulate data volume. Samples passing quality filtering are directly added to the data pool $\mathcal{D}_{t}$ without any density constraint.
\textbf{Density fine-tuning phase} ($T \rightarrow T^{\text{target}}$): As the token count approaches $T^{\text{target}}$, the process transitions to density fine-tuning, with knowledge density as the control target. The acceptance strategy is as follows: using the density formula in Sec. \ref{Knowledge Density Definition}, we back-calculate $r^{\text{target}}$ from $T^{\text{target}}$ and $\rho^{\text{target}}$. If the current $r < r^{\text{target}}$ (density too high), we preferentially accept questions far from the centroid to increase $r$; if $r > r^{\text{target}}$ (density too low), we preferentially accept questions close to the centroid to decrease $r$.
Stages 2.1\textasciitilde2.3 iterate until the data pool satisfies the convergence condition (We set the maximum number of iterations to $k$.):

\vspace{-10pt}
\begin{equation}
\scalebox{0.9}{$\displaystyle
\left| T - T^{target} \middle| < \epsilon^{T}, \middle| \rho - \rho^{target} \middle| < \epsilon^{\rho} \right.
$}
\end{equation}

Finally, we obtain the data pool conforming to the target distribution $P(\mathcal{S}^{\text{syn}}) \sim \mathcal{D}^{\text{target}}$. The seed question pool of 14M is expanded to 71M synthetic samples. The detailed rejection sampling algorithm is provided in Appendix \ref{Algorithm Details}.

\section{Experiment}

\subsection{Experimental Settings}

\paragraph{Datasets and Tasks.}
We evaluate on six knowledge benchmarks, divided into knowledge-intensive sets: \textbf{Web Questions (WebQ)}, \textbf{Natural Questions (NQ)}, and \textbf{TriviaQA (TriQA)}; and long-tail knowledge sets: \textbf{SimpleQA (Sim)}, \textbf{SimpleQA-Verified (Sim-V)}, and \textbf{EntityQuestions (EQ)}. Details of benchmarks are provided in Appendix \ref{Dataset Statistics}.

\paragraph{Baselines.}
We compare four synthesis strategies: \textbf{Rand.} (Random) applies no control and stops at target tokens; \textbf{Uni.} (Uniform) enforces equal ratios across domains; \textbf{Diff.} (Difficulty-weighted Importance Synthesis) prioritizes high-PPL samples to learn harder knowledge first; \textbf{Qual.} (Quality-filtered Rejection Synthesis) prioritizes high-quality samples from LLM judges.

\paragraph{Evaluation Metrics.}
We use accuracy and cross-entropy loss as evaluation metrics. For accuracy, we adopt an LLM-as-judge approach, using Qwen3.5-397B-A17B to classify each model prediction as Correct, Incorrect, or Not Attempted; accuracy is defined as the fraction of Correct samples. Cross-entropy loss measures the model's tendency to generate the correct answer, computed over the gold answer tokens.

\paragraph{Implementation Details.}
We collect approximately 14M seed QA pairs (1.73B tokens) from Wikipedia, which KDoS expands to 71M samples (9.28B tokens), with a maximum iteration count of $k=200$. Data synthesis and quality filtering are conducted on an NVIDIA H20-3E cluster with 128 nodes $\times$ 8 GPUs (1024 H20-3E GPUs in total). Knowledge injection experiments are conducted on models including Qwen3.0-base \citep{DBLP:journals/corr/abs-2505-09388}, Ling-mini-2.0-base \citep{team2025every}, and LLaMA-3.2-base \citep{grattafiori2024llama}, with Qwen3-4B-Base as the default backbone, using an NVIDIA H800 cluster with 8 nodes $\times$ 8 GPUs (64 H800 GPUs in total). More details are provided in Appendix \ref{Supplement Implementation Details}.

\subsection{Main Result}


\begin{table}[ht]
\centering
\resizebox{1.0\columnwidth}{!}{%
\begin{tabular}{lccccccc}
\toprule
Method & \textbf{WebQ} & \textbf{TriQA} & \textbf{NQ} & \textbf{Sim} & \textbf{Sim-V} & \textbf{EQ} & \cellcolor[HTML]{E8F7CF}\textbf{Avg.} \\ \midrule
Base & 18.7 & 32.9 & 11.9 & 3.5 & 3.6 & 10.2 & \cellcolor[HTML]{E8F7CF}16.1 \\
SP & 23.0 & 33.4 & 13.3 & 5.8 & 6.8 & 11.1 & \cellcolor[HTML]{E8F7CF}17.4 \\ \midrule
\rowcolor{gray!20}
\multicolumn{8}{c}{\textbf{\textit{Synthesis Strategy}}} \\ \midrule
Rand. & 26.1 & 32.8 & 18.0 & 6.0 & 6.7 & 12.1 & \cellcolor[HTML]{E8F7CF}18.3 \\
Uni. & {\ul 27.9} & 37.3 & 19.7 & 6.6 & 7.9 & 14.6 & \cellcolor[HTML]{E8F7CF}20.9 \\
Qual. & 26.9 & 36.3 & 19.8 & 6.3 & 6.8 & 14.8 & \cellcolor[HTML]{E8F7CF}20.6 \\
Diff. & 26.0 & {\ul 38.5} & {\ul 19.9} & {\ul 7.2} & \textbf{8.2} & {\ul 15.2} & \cellcolor[HTML]{E8F7CF}{\ul 21.5} \\ \midrule
\textbf{KDoS} & \textbf{31.8} & \textbf{39.3} & \textbf{21.6} & \textbf{8.0} & {\ul 8.0} & \textbf{16.4} & \cellcolor[HTML]{E8F7CF}\textbf{22.8} \\ \bottomrule
\end{tabular}
}
\caption{Performance comparison of KDoS and other synthesis strategies on six knowledge benchmarks, using Qwen3-4B-Base as the backbone. SP denotes Seed Pool. \textbf{Bold} and \underline{underline} indicate the best and second-best results, respectively.}
\label{tab:main-results}
\end{table}


As shown in Tab.~\ref{tab:main-results}, the base model achieves an average score of 16.1 across six benchmarks. Training with the seed pool (SP) increases the average by 1.3 points. Among synthesis methods, Rand. achieves 18.3 but underperforms SP on some benchmarks, indicating that blind synthesis leads to uncontrolled distribution where some knowledge points are redundantly repeated while others remain critically sparse, limiting LLM knowledge boundaries. Uni. improves to 20.9 by balancing domain ratios. Qual. achieves 20.6 and Diff. achieves 21.5, ranking second overall. In contrast, KDoS converges to the optimal knowledge distribution, achieving the best average of 22.8—1.3 points above Diff. and 1.9 points above Uni.—demonstrating the effectiveness of distribution-optimized synthesis.

\subsection{Ablation Study}


\begin{table}[ht]
\centering
\resizebox{0.8\columnwidth}{!}{%
\begin{tabular}{lcccc}
\toprule
\multirow{2}{*}{Benchmark} & \multicolumn{4}{c}{Module} \\
\cmidrule(lr){2-5}
& \textbf{KDoS} & w/o F & w/o F \& D & SP \\ \midrule
EQ & 16.4 & 15.5 & 11.5 & 11.1 \\
Sim & 8.0 & 7.0 & 5.8 & 5.8 \\
Simp-V & 8.0 & 7.2 & 6.6 & 6.8 \\
WebQ & 31.8 & 29.0 & 23.4 & 23.0 \\
NQ & 21.6 & 20.1 & 13.8 & 13.3 \\
TriQA & 39.3 & 38.4 & 33.5 & 33.4 \\ \midrule
\rowcolor[HTML]{E8F7CF} 
Average & \textbf{22.8} & 21.7 & 17.6 & 17.4 \\ \bottomrule
\end{tabular}
}
\caption{Ablation study of KDoS modules.} 
\label{tab:ablation}
\end{table}

We evaluate the effectiveness of the Quality Filtering (F) and Density-aware Rejection Sampling (D) modules. We compare four configurations: (1) full KDoS, (2) w/o F, (3) w/o F \& D (i.e., only deduplication on the seed pool), and (4) the seed pool (SP) baseline. As shown in Tab.~\ref{tab:ablation}, removing F leads to a 1.07\% drop in average score, while removing both F and D results in a 5.19\% drop, validating the effectiveness of both modules. The results also indicate that distribution optimization plays the dominant role in performance improvement.

\subsection{Scaling with Model and Data Size}
\label{scaling with model and data size}

\begin{figure*}[ht]
    \centering
    \includegraphics[width=1.0\textwidth, keepaspectratio]{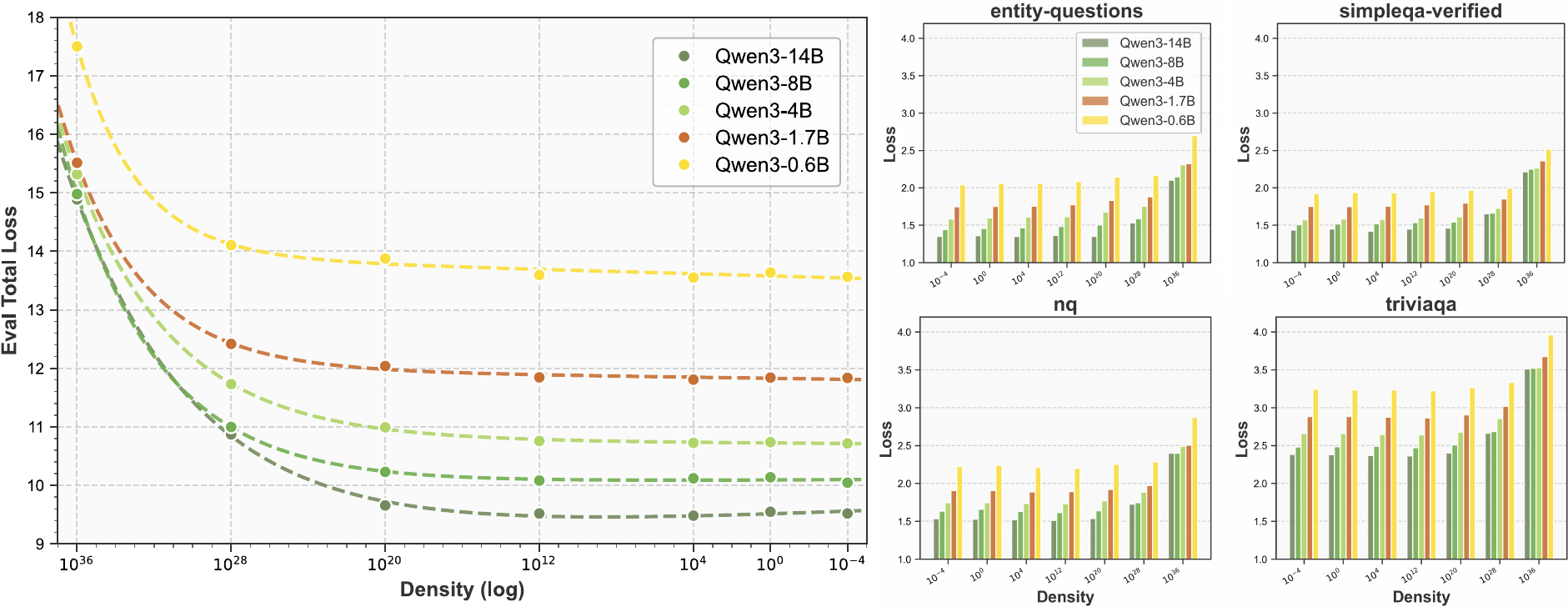}
    \caption{Scaling with model size. \textbf{Left:} Total eval loss scaling curves of Qwen3-base models of different sizes trained on synthetic data with varying densities. \textbf{Right:} Per-dataset loss on EQ, SimpleQA-V, NQ, and TriviaQA.}
    \label{fig:scaling_model_size}
\end{figure*}

\begin{figure*}[ht]
    \centering
    \includegraphics[width=1.0\textwidth, keepaspectratio]{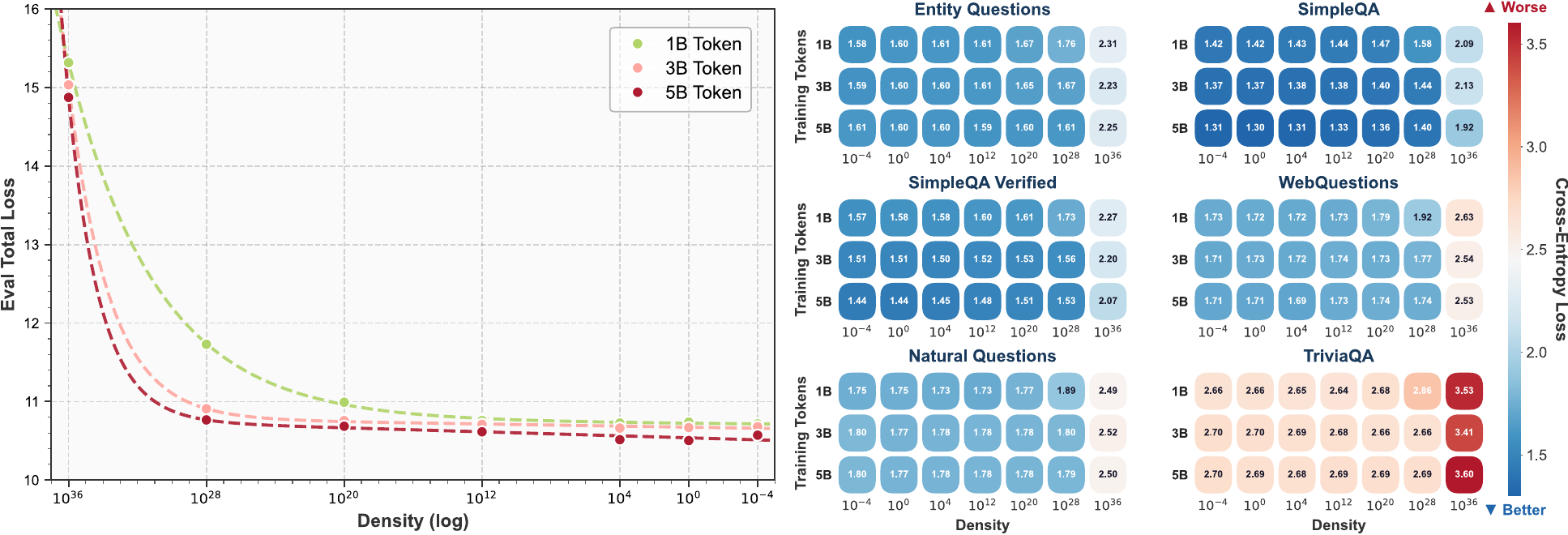}
    \caption{Scaling with data size. \textbf{Left:} Total eval loss scaling curves of Qwen3-4B-Base trained on synthetic data with varying densities and data sizes. \textbf{Right:} Heatmap of data size $\times$ density $\times$ loss.}
    \label{fig:scaling_data_size}
\end{figure*}

We investigate the effect of knowledge density on knowledge injection across Qwen3-base models of varying sizes (0.6B$\sim$14B) and data volumes (1B, 3B, 5B tokens). As shown in Fig.~\ref{fig:scaling_model_size} and Fig.~\ref{fig:scaling_data_size}, all settings exhibit stable and consistent scaling curves, with the lowest eval total loss consistently achieved in the density range of $10^{-4}$\textasciitilde$10^{4}$. As model size or data volume increases, the curves shift downward consistently. These results demonstrate that an optimal knowledge density range exists across all model and data scales, maximizing knowledge boundary scaling efficiency.
All settings also exhibit a \textit{Knowledge Collapse Region} where loss increases sharply at excessively high densities. Notably, larger models reach the \textit{Knowledge Collapse Point} at lower densities (e.g., $10^{28}$ for 0.6B vs. $10^{12}$ for 14B), suggesting that larger models require finer-grained distribution control. Meanwhile, larger data volumes reach the \textit{Knowledge Collapse Point} at higher densities (e.g., $10^{12}$ for 1B tokens vs. $10^{28}$ for 5B tokens), suggesting that different data scales require different levels of distribution control granularity. This may further indicate that the optimal density range varies across training stages: pre-training with larger data volumes may tolerate a wider optimal density range than post-training, likely because more data increases the absolute count of knowledge points in high-density regions, enabling the model to maintain learning efficiency in denser knowledge environments.

\subsection{Scaling with LLM Backbones}

\begin{figure*}[ht]
    \centering
    \includegraphics[width=1.0\textwidth, keepaspectratio]{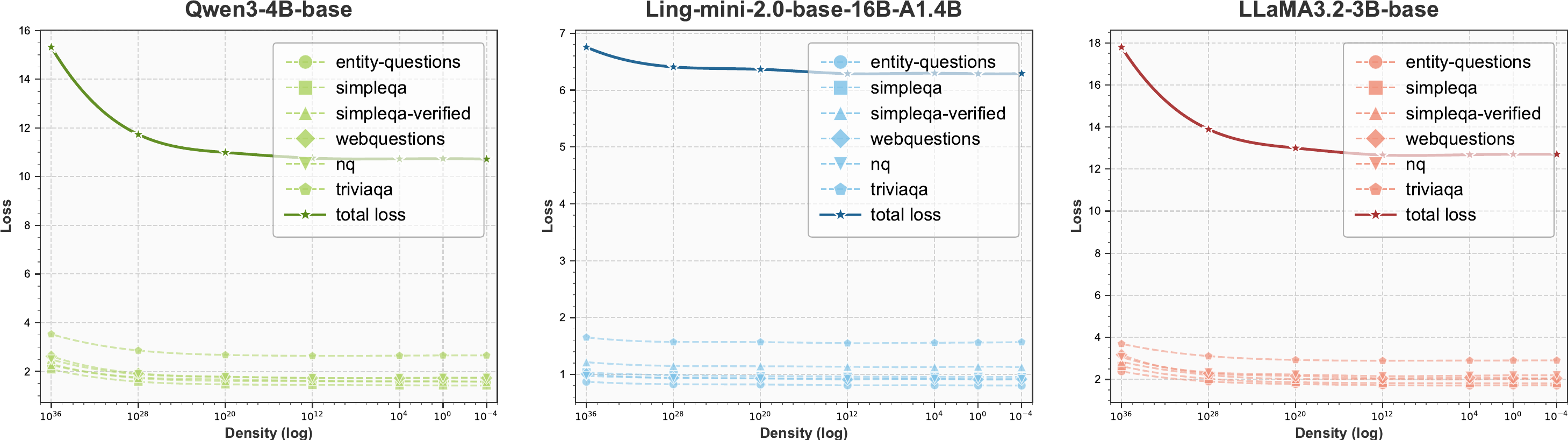} 
    \caption{Scaling with differenct LLM backbones.}  
    \label{fig:backbone}
\end{figure*}

We investigate the effect of knowledge density on knowledge injection across different backbone LLMs, including Qwen3-4B-Base, Ling-mini-2.0-16B-A3B, and LLaMA-3.2-3B. As shown in Fig.~\ref{fig:backbone}, regardless of whether the model is a dense or Mixture-of-Experts (MoE) architecture, all backbones exhibit stable and consistent scaling curves, with the lowest eval total loss consistently achieved in the density range of $10^{-4}$\textasciitilde$10^{4}$. This is consistent with the findings in Sec.~\ref{scaling with model and data size}. 
\begin{Promptbox}{Scaling Principle of Knowledge Distribution}
Across varying model sizes, data scales, and backbone architectures, an optimal knowledge density range consistently exists, which on our data falls within $10^{-4}$\textasciitilde$10^{4}$, revealing a general and robust principle of knowledge injection.
\end{Promptbox}

\subsection{Efficiency of Knowledge Injection}

\begin{figure*}[ht]
    \centering
    \includegraphics[width=1.0\textwidth, keepaspectratio]{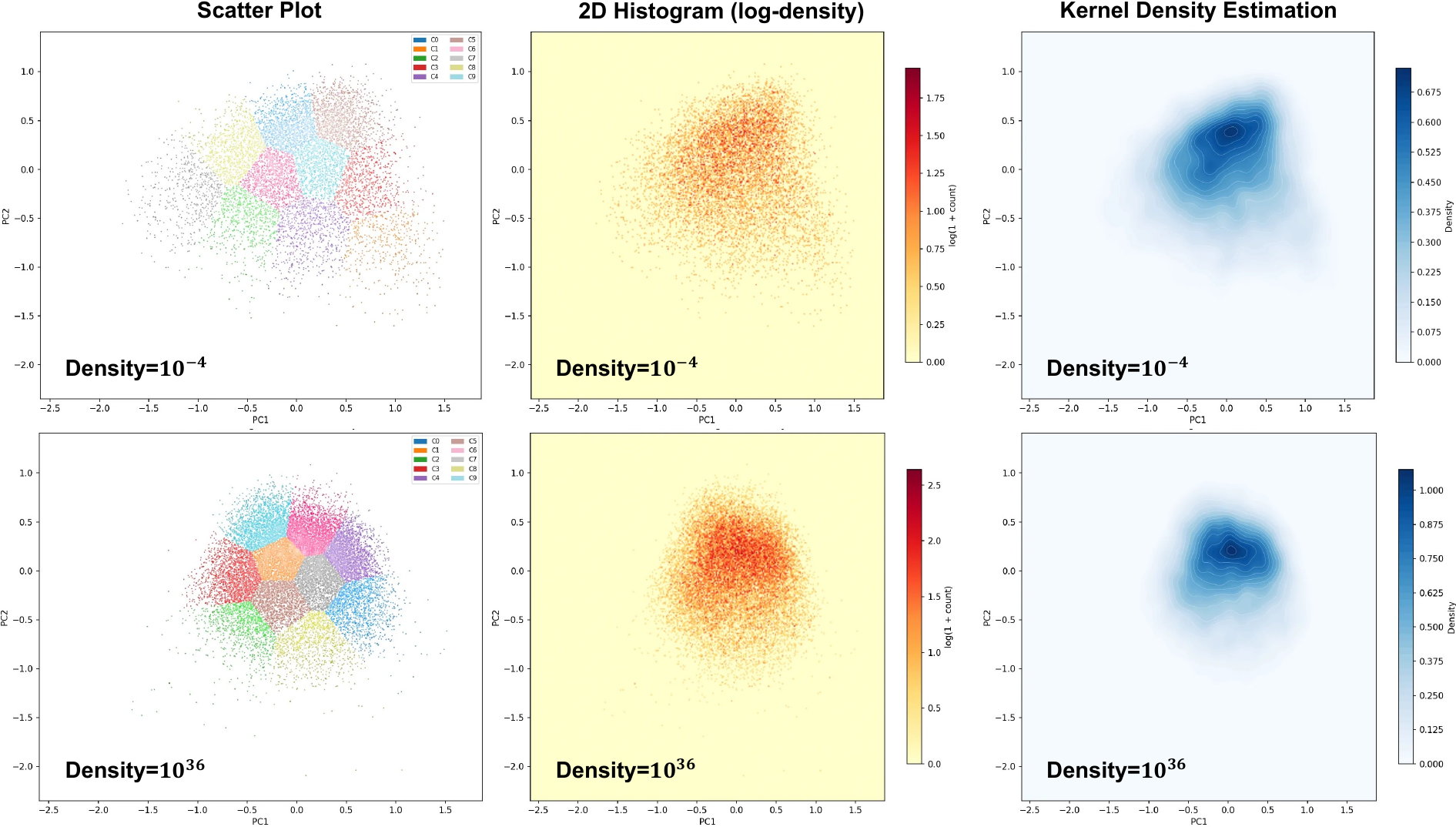} 
    \caption{Case study comparing data with different knowledge densities.}  
    \label{fig:case_study}
\end{figure*}

\begin{figure}[ht]
    \centering
    \includegraphics[width=0.6\columnwidth, keepaspectratio]{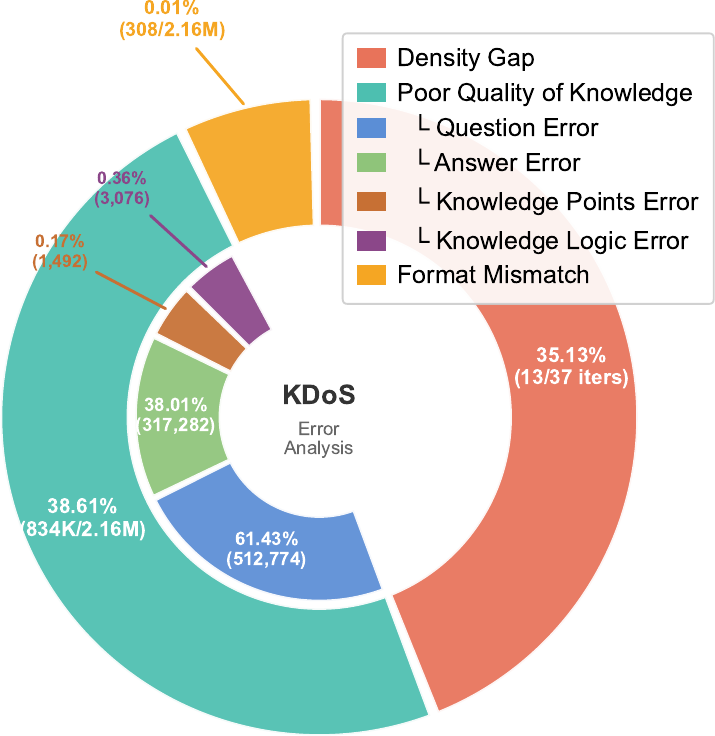}
    \caption{Error analysis of KDoS.}
    \label{fig:error_analysis}
\end{figure}

\begin{figure*}[ht]
    \centering
    \includegraphics[width=1.0\textwidth, keepaspectratio]{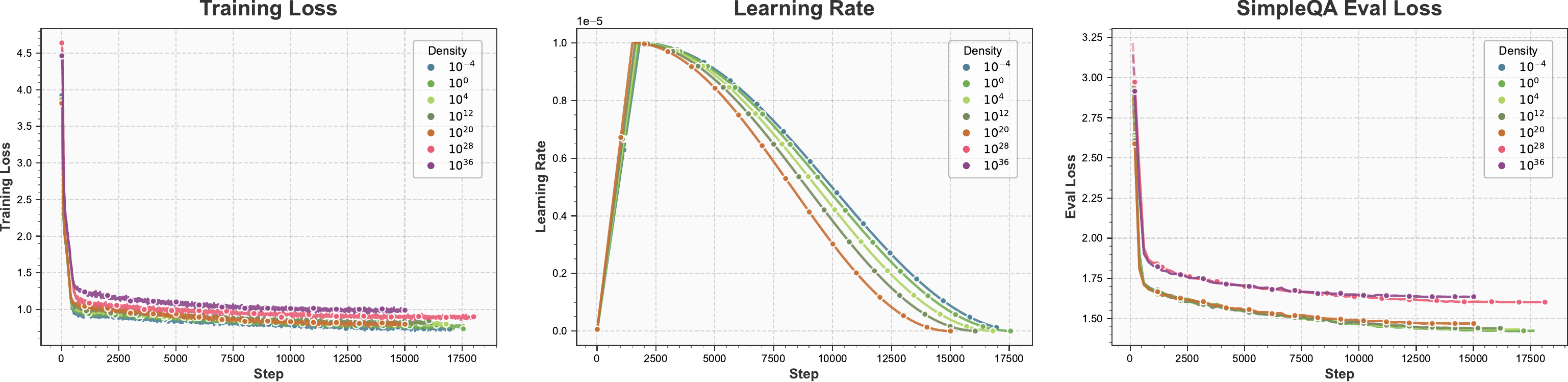} 
    \caption{Efficiency of Knowledge Injection. We compare the training loss, learning rate, and SimpleQA test set eval loss curves of synthetic data with different knowledge densities throughout training.}  
    \label{fig:efficiency}
\end{figure*}

We analyze knowledge injection efficiency across different knowledge distributions by examining training loss, learning rate, and eval loss. As shown in Fig.~\ref{fig:efficiency}, higher density consistently leads to higher converged training loss, lower learning rate at the same training step, and higher converged eval loss. This indicates that higher-density knowledge distributions generally result in lower injection efficiency, while distributions within the optimal density range ($10^{-4}$\textasciitilde$10^{4}$) consistently maintain high injection efficiency.

\subsection{Error Analysis}

We categorize errors in KDoS into three types: (1) \textbf{Density Gap}, the number of iterations where the density converges in the wrong direction; (2) \textbf{Poor Knowledge Quality}, various quality defects in synthesized samples, further divided into (2.1) Question Error, (2.2) Answer Error, (2.3) Knowledge Points Error, and (2.4) Knowledge Logic Error; and (3) \textbf{Format Mismatch}, incorrect LLM output formats. \textbf{Note that error rates across the three types are not directly comparable.}
We report error statistics from a single run that expands 1.73B seed data to 2B (approximately 2.16M new samples). Over 37 iterations, 13 Density Gap errors occurred. Among the 2,161,357 synthesized samples, 834,624 exhibited type (2) errors and 308 exhibited type (1) errors. Within type (2), Question Error (2.1) and Answer Error (2.2) are the dominant subtypes, as shown in Fig.~\ref{fig:error_analysis}.

\subsection{Case Study}
\label{Case Study}

We visualize the knowledge distributions at different density levels, as shown in Fig.~\ref{fig:case_study} (full visualizations in Appendix \ref{Complete Case Study}). We compare scatter plots, 2D histograms, and kernel density estimation (KDE) across different distributions. Visually, higher-density distributions occupy a smaller average radius in semantic space, indicating more concentrated knowledge coverage.

\section{Conclusion}

This paper proposes Knowledge Distribution-Optimized Synthesis, which improves LLM knowledge injection efficiency from the knowledge distribution optimization perspective. We introduce knowledge density as a controllable variable and employ a three-stage dynamic feedback mechanism to precisely shape the knowledge distribution of synthetic data, shifting the paradigm from blind synthesis to distribution-driven synthesis.
Our experiments reveal a stable optimal knowledge density range that maximizes knowledge boundary expansion across different model scales (0.6B\textasciitilde16B) and data scales (1B\textasciitilde5B tokens), demonstrating notable stability and highlighting a general principle in knowledge injection. Extensive experiments demonstrate that KDoS scales LLM knowledge boundaries through precise distribution control, significantly outperforming existing baselines across six established knowledge benchmarks.

\section*{Limitations}

To the best of our knowledge, our method primarily contains the following limitation: 

Our work focuses on the post-training stage, specifically the SFT phase, to explore the scaling of LLM knowledge boundaries and derives general laws governing knowledge injection through optimized data distribution. However, we do not extend our investigation to broader settings such as the pre-training stage, where analogous knowledge distribution optimization may also yield meaningful gains. This is primarily due to the substantially larger data volumes and computational overhead required for pre-training experiments.

\bibliography{custom}

@article{zhang2025adept,
  title={ADEPT: Continual Pretraining via Adaptive Expansion and Dynamic Decoupled Tuning},
  author={Zhang, Jinyang and Fang, Yue and Ding, Hongxin and Liao, Weibin and Ye, Muyang and Chu, Xu and Zhao, Junfeng and Wang, Yasha},
  journal={arXiv preprint arXiv:2510.10071},
  year={2025}
}

@inproceedings{10.5555/3618408.3619049,
author = {Kandpal, Nikhil and Deng, Haikang and Roberts, Adam and Wallace, Eric and Raffel, Colin},
title = {Large language models struggle to learn long-tail knowledge},
year = {2023},
publisher = {JMLR.org},
booktitle = {Proceedings of the 40th International Conference on Machine Learning},
articleno = {641},
numpages = {12},
location = {Honolulu, Hawaii, USA},
series = {ICML'23}
}

@article{haas2025simpleqa,
  title={Simpleqa verified: A reliable factuality benchmark to measure parametric knowledge},
  author={Haas, Lukas and Yona, Gal and D'Antonio, Giovanni and Goldshtein, Sasha and Das, Dipanjan},
  journal={arXiv preprint arXiv:2509.07968},
  year={2025}
}

@article{wei2024measuring,
  title={Measuring short-form factuality in large language models},
  author={Wei, Jason and Karina, Nguyen and Chung, Hyung Won and Jiao, Yunxin Joy and Papay, Spencer and Glaese, Amelia and Schulman, John and Fedus, William},
  journal={arXiv preprint arXiv:2411.04368},
  year={2024}
}

@inproceedings{sciavolino2021simple,
   title={Simple Entity-centric Questions Challenge Dense Retrievers},
   author={Sciavolino, Christopher and Zhong, Zexuan and Lee, Jinhyuk and Chen, Danqi},
   booktitle={Empirical Methods in Natural Language Processing (EMNLP)},
   year={2021}
}

@inproceedings{lv2025inject,
  title={How to inject knowledge efficiently? Knowledge Infusion Scaling Law for Pre-training Large Language Models},
  author={Lv, Kangtao and Chen, Haibin and Yuan, Yujin and Liu, Langming and Liu, Shilei and Wang, Yongwei and Su, Wenbo and Zheng, Bo},
  booktitle={Proceedings of the 2025 Conference on Empirical Methods in Natural Language Processing},
  pages={26204--26219},
  year={2025}
}

@inproceedings{ovadia2024fine,
  title={Fine-tuning or retrieval? comparing knowledge injection in llms},
  author={Ovadia, Oded and Brief, Menachem and Mishaeli, Moshik and Elisha, Oren},
  booktitle={Proceedings of the 2024 conference on empirical methods in natural language processing},
  pages={237--250},
  year={2024}
}

@inproceedings{joshi2017triviaqa,
  title={Triviaqa: A large scale distantly supervised challenge dataset for reading comprehension},
  author={Joshi, Mandar and Choi, Eunsol and Weld, Daniel S and Zettlemoyer, Luke},
  booktitle={Proceedings of the 55th Annual Meeting of the Association for Computational Linguistics (Volume 1: Long Papers)},
  pages={1601--1611},
  year={2017}
}

@article{kwiatkowski2019natural,
  title={Natural questions: a benchmark for question answering research},
  author={Kwiatkowski, Tom and Palomaki, Jennimaria and Redfield, Olivia and Collins, Michael and Parikh, Ankur and Alberti, Chris and Epstein, Danielle and Polosukhin, Illia and Devlin, Jacob and Lee, Kenton and others},
  journal={Transactions of the Association for Computational Linguistics},
  volume={7},
  pages={453--466},
  year={2019},
  publisher={MIT Press One Rogers Street, Cambridge, MA 02142-1209, USA journals-info~…}
}

@misc{talmor2018webknowledgebaseansweringcomplex,
      title={The Web as a Knowledge-base for Answering Complex Questions}, 
      author={Alon Talmor and Jonathan Berant},
      year={2018},
      eprint={1803.06643},
      archivePrefix={arXiv},
      primaryClass={cs.CL},
      url={https://arxiv.org/abs/1803.06643}, 
}

@article{meng2022locating,
  title={Locating and editing factual associations in gpt},
  author={Meng, Kevin and Bau, David and Andonian, Alex and Belinkov, Yonatan},
  journal={Advances in neural information processing systems},
  volume={35},
  pages={17359--17372},
  year={2022}
}

@article{meng2022mass,
  title={Mass-editing memory in a transformer},
  author={Meng, Kevin and Sharma, Arnab Sen and Andonian, Alex and Belinkov, Yonatan and Bau, David},
  journal={arXiv preprint arXiv:2210.07229},
  year={2022}
}

@inproceedings{wu2024llama,
  title={Llama pro: Progressive llama with block expansion},
  author={Wu, Chengyue and Gan, Yukang and Ge, Yixiao and Lu, Zeyu and Wang, Jiahao and Feng, Ye and Shan, Ying and Luo, Ping},
  booktitle={Proceedings of the 62nd Annual Meeting of the Association for Computational Linguistics (Volume 1: Long Papers)},
  pages={6518--6537},
  year={2024}
}

@inproceedings{wang2023self,
  title={Self-instruct: Aligning language models with self-generated instructions},
  author={Wang, Yizhong and Kordi, Yeganeh and Mishra, Swaroop and Liu, Alisa and Smith, Noah A and Khashabi, Daniel and Hajishirzi, Hannaneh},
  booktitle={Proceedings of the 61st annual meeting of the association for computational linguistics (volume 1: long papers)},
  pages={13484--13508},
  year={2023}
}

@article{xu2023wizardlm,
  title={Wizardlm: Empowering large language models to follow complex instructions},
  author={Xu, Can and Sun, Qingfeng and Zheng, Kai and Geng, Xiubo and Zhao, Pu and Feng, Jiazhan and Tao, Chongyang and Jiang, Daxin},
  journal={arXiv preprint arXiv:2304.12244},
  year={2023}
}

@inproceedings{xu2025magpie,
  title={Magpie: Alignment data synthesis from scratch by prompting aligned llms with nothing},
  author={Xu, Zhangchen and Jiang, Fengqing and Niu, Luyao and Deng, Yuntian and Poovendran, Radha and Choi, Yejin and Lin, Bill Yuchen},
  booktitle={International Conference on Learning Representations},
  volume={2025},
  pages={76346--76382},
  year={2025}
}

@misc{zelikman2022starbootstrappingreasoningreasoning,
      title={STaR: Bootstrapping Reasoning With Reasoning}, 
      author={Eric Zelikman and Yuhuai Wu and Jesse Mu and Noah D. Goodman},
      year={2022},
      eprint={2203.14465},
      archivePrefix={arXiv},
      primaryClass={cs.LG},
      url={https://arxiv.org/abs/2203.14465}, 
}

@misc{yuan2023scalingrelationshiplearningmathematical,
      title={Scaling Relationship on Learning Mathematical Reasoning with Large Language Models}, 
      author={Zheng Yuan and Hongyi Yuan and Chengpeng Li and Guanting Dong and Keming Lu and Chuanqi Tan and Chang Zhou and Jingren Zhou},
      year={2023},
      eprint={2308.01825},
      archivePrefix={arXiv},
      primaryClass={cs.CL},
      url={https://arxiv.org/abs/2308.01825}, 
}

@misc{tong2024dartmathdifficultyawarerejectiontuning,
      title={DART-Math: Difficulty-Aware Rejection Tuning for Mathematical Problem-Solving}, 
      author={Yuxuan Tong and Xiwen Zhang and Rui Wang and Ruidong Wu and Junxian He},
      year={2024},
      eprint={2407.13690},
      archivePrefix={arXiv},
      primaryClass={cs.CL},
      url={https://arxiv.org/abs/2407.13690}, 
}

@misc{liu2024makesgooddataalignment,
      title={What Makes Good Data for Alignment? A Comprehensive Study of Automatic Data Selection in Instruction Tuning}, 
      author={Wei Liu and Weihao Zeng and Keqing He and Yong Jiang and Junxian He},
      year={2024},
      eprint={2312.15685},
      archivePrefix={arXiv},
      primaryClass={cs.CL},
      url={https://arxiv.org/abs/2312.15685}, 
}

@misc{wang2025treesynthsynthesizingdiversedata,
      title={TreeSynth: Synthesizing Diverse Data from Scratch via Tree-Guided Subspace Partitioning}, 
      author={Sheng Wang and Pengan Chen and Jingqi Zhou and Qintong Li and Jingwei Dong and Jiahui Gao and Boyang Xue and Jiyue Jiang and Lingpeng Kong and Chuan Wu},
      year={2025},
      eprint={2503.17195},
      archivePrefix={arXiv},
      primaryClass={cs.LG},
      url={https://arxiv.org/abs/2503.17195}, 
}

@misc{chen2025graphgenenhancingsupervisedfinetuning,
      title={GraphGen: Enhancing Supervised Fine-Tuning for LLMs with Knowledge-Driven Synthetic Data Generation}, 
      author={Zihong Chen and Wanli Jiang and Jinzhe Li and Zhonghang Yuan and Huanjun Kong and Wanli Ouyang and Nanqing Dong},
      year={2025},
      eprint={2505.20416},
      archivePrefix={arXiv},
      primaryClass={cs.CL},
      url={https://arxiv.org/abs/2505.20416}, 
}

@inproceedings{wang-etal-2024-codeclm,
    title = "{C}odec{LM}: Aligning Language Models with Tailored Synthetic Data",
    author = "Wang, Zifeng  and
      Li, Chun-Liang  and
      Perot, Vincent  and
      Le, Long  and
      Miao, Jin  and
      Zhang, Zizhao  and
      Lee, Chen-Yu  and
      Pfister, Tomas",
    editor = "Duh, Kevin  and
      Gomez, Helena  and
      Bethard, Steven",
    booktitle = "Findings of the Association for Computational Linguistics: NAACL 2024",
    month = jun,
    year = "2024",
    address = "Mexico City, Mexico",
    publisher = "Association for Computational Linguistics",
    url = "https://aclanthology.org/2024.findings-naacl.235/",
    doi = "10.18653/v1/2024.findings-naacl.235",
    pages = "3712--3729"
}

@article{Qin2025ScalingLO,
  title={Scaling Laws of Synthetic Data for Language Models},
  author={Zeyu Qin and Qingxiu Dong and Xingxing Zhang and Li Dong and Xiaolong Huang and Ziyi Yang and Mahmoud Khademi and Dongdong Zhang and Hany Hassan Awadalla and Yi R. Fung and Weizhu Chen and Minhao Cheng and Furu Wei},
  journal={ArXiv},
  year={2025},
  volume={abs/2503.19551},
  url={https://api.semanticscholar.org/CorpusID:277313659}
}

@inproceedings{chen-etal-2025-revisiting,
    title = "Revisiting Scaling Laws for Language Models: The Role of Data Quality and Training Strategies",
    author = "Chen, Zhengyu  and
      Wang, Siqi  and
      Xiao, Teng  and
      Wang, Yudong  and
      Chen, Shiqi  and
      Cai, Xunliang  and
      He, Junxian  and
      Wang, Jingang",
    editor = "Che, Wanxiang  and
      Nabende, Joyce  and
      Shutova, Ekaterina  and
      Pilehvar, Mohammad Taher",
    booktitle = "Proceedings of the 63rd Annual Meeting of the Association for Computational Linguistics (Volume 1: Long Papers)",
    month = jul,
    year = "2025",
    address = "Vienna, Austria",
    publisher = "Association for Computational Linguistics",
    url = "https://aclanthology.org/2025.acl-long.1163/",
    doi = "10.18653/v1/2025.acl-long.1163",
    pages = "23881--23899",
    ISBN = "979-8-89176-251-0"
}

@inproceedings{li-etal-2025-knowledge-boundary,
    title = "Knowledge Boundary of Large Language Models: A Survey",
    author = "Li, Moxin  and
      Zhao, Yong  and
      Zhang, Wenxuan  and
      Li, Shuaiyi  and
      Xie, Wenya  and
      Ng, See-Kiong  and
      Chua, Tat-Seng  and
      Deng, Yang",
    editor = "Che, Wanxiang  and
      Nabende, Joyce  and
      Shutova, Ekaterina  and
      Pilehvar, Mohammad Taher",
    booktitle = "Proceedings of the 63rd Annual Meeting of the Association for Computational Linguistics (Volume 1: Long Papers)",
    month = jul,
    year = "2025",
    address = "Vienna, Austria",
    publisher = "Association for Computational Linguistics",
    url = "https://aclanthology.org/2025.acl-long.256/",
    doi = "10.18653/v1/2025.acl-long.256",
    pages = "5131--5157",
    ISBN = "979-8-89176-251-0"
}

@inproceedings{sun-etal-2024-head,
    title = "Head-to-Tail: How Knowledgeable are Large Language Models ({LLM}s)? {A}.{K}.{A}. Will {LLM}s Replace Knowledge Graphs?",
    author = "Sun, Kai  and
      Xu, Yifan  and
      Zha, Hanwen  and
      Liu, Yue  and
      Dong, Xin Luna",
    editor = "Duh, Kevin  and
      Gomez, Helena  and
      Bethard, Steven",
    booktitle = "Proceedings of the 2024 Conference of the North American Chapter of the Association for Computational Linguistics: Human Language Technologies (Volume 1: Long Papers)",
    month = jun,
    year = "2024",
    address = "Mexico City, Mexico",
    publisher = "Association for Computational Linguistics",
    url = "https://aclanthology.org/2024.naacl-long.18/",
    doi = "10.18653/v1/2024.naacl-long.18",
    pages = "311--325"
}

@inproceedings{zhao-etal-2025-know,
    title = "Do We Know What {LLM}s Don{'}t Know? A Study of Consistency in Knowledge Probing",
    author = {Zhao, Raoyuan  and
      K{\"o}ksal, Abdullatif  and
      Modarressi, Ali  and
      Hedderich, Michael A.  and
      Schuetze, Hinrich},
    editor = "Christodoulopoulos, Christos  and
      Chakraborty, Tanmoy  and
      Rose, Carolyn  and
      Peng, Violet",
    booktitle = "Findings of the Association for Computational Linguistics: EMNLP 2025",
    month = nov,
    year = "2025",
    address = "Suzhou, China",
    publisher = "Association for Computational Linguistics",
    url = "https://aclanthology.org/2025.findings-emnlp.1263/",
    doi = "10.18653/v1/2025.findings-emnlp.1263",
    pages = "23254--23280",
    ISBN = "979-8-89176-335-7"
}

@misc{yin2023largelanguagemodelsknow,
      title={Do Large Language Models Know What They Don't Know?}, 
      author={Zhangyue Yin and Qiushi Sun and Qipeng Guo and Jiawen Wu and Xipeng Qiu and Xuanjing Huang},
      year={2023},
      eprint={2305.18153},
      archivePrefix={arXiv},
      primaryClass={cs.CL},
      url={https://arxiv.org/abs/2305.18153}, 
}

@inproceedings{mallen-etal-2023-trust,
    title = "When Not to Trust Language Models: Investigating Effectiveness of Parametric and Non-Parametric Memories",
    author = "Mallen, Alex  and
      Asai, Akari  and
      Zhong, Victor  and
      Das, Rajarshi  and
      Khashabi, Daniel  and
      Hajishirzi, Hannaneh",
    editor = "Rogers, Anna  and
      Boyd-Graber, Jordan  and
      Okazaki, Naoaki",
    booktitle = "Proceedings of the 61st Annual Meeting of the Association for Computational Linguistics (Volume 1: Long Papers)",
    month = jul,
    year = "2023",
    address = "Toronto, Canada",
    publisher = "Association for Computational Linguistics",
    url = "https://aclanthology.org/2023.acl-long.546/",
    doi = "10.18653/v1/2023.acl-long.546",
    pages = "9802--9822"
}

@inproceedings{10.5555/3524938.3525306,
author = {Guu, Kelvin and Lee, Kenton and Tung, Zora and Pasupat, Panupong and Chang, Ming-Wei},
title = {REALM: retrieval-augmented language model pre-training},
year = {2020},
publisher = {JMLR.org},
booktitle = {Proceedings of the 37th International Conference on Machine Learning},
articleno = {368},
numpages = {10},
series = {ICML'20}
}

@misc{allenzhu2024physicslanguagemodels33,
      title={Physics of Language Models: Part 3.3, Knowledge Capacity Scaling Laws}, 
      author={Zeyuan Allen-Zhu and Yuanzhi Li},
      year={2024},
      eprint={2404.05405},
      archivePrefix={arXiv},
      primaryClass={cs.CL},
      url={https://arxiv.org/abs/2404.05405}, 
}

@misc{ke2023continualpretraininglanguagemodels,
      title={Continual Pre-training of Language Models}, 
      author={Zixuan Ke and Yijia Shao and Haowei Lin and Tatsuya Konishi and Gyuhak Kim and Bing Liu},
      year={2023},
      eprint={2302.03241},
      archivePrefix={arXiv},
      primaryClass={cs.CL},
      url={https://arxiv.org/abs/2302.03241}, 
}

@inproceedings{10.5555/3666122.3666237,
author = {Sun, Zhiqing and Shen, Yikang and Zhou, Qinhong and Zhang, Hongxin and Chen, Zhenfang and Cox, David and Yang, Yiming and Gan, Chuang},
title = {Principle-driven self-alignment of language models from scratch with minimal human supervision},
year = {2023},
publisher = {Curran Associates Inc.},
address = {Red Hook, NY, USA},
booktitle = {Proceedings of the 37th International Conference on Neural Information Processing Systems},
articleno = {115},
numpages = {55},
location = {New Orleans, LA, USA},
series = {NIPS '23}
}

@misc{yang2024syntheticcontinuedpretraining,
      title={Synthetic continued pretraining}, 
      author={Zitong Yang and Neil Band and Shuangping Li and Emmanuel Candès and Tatsunori Hashimoto},
      year={2024},
      eprint={2409.07431},
      archivePrefix={arXiv},
      primaryClass={cs.LG},
      url={https://arxiv.org/abs/2409.07431}, 
}

@misc{azerbayev2024llemmaopenlanguagemodel,
      title={Llemma: An Open Language Model For Mathematics}, 
      author={Zhangir Azerbayev and Hailey Schoelkopf and Keiran Paster and Marco Dos Santos and Stephen McAleer and Albert Q. Jiang and Jia Deng and Stella Biderman and Sean Welleck},
      year={2024},
      eprint={2310.10631},
      archivePrefix={arXiv},
      primaryClass={cs.CL},
      url={https://arxiv.org/abs/2310.10631}, 
}

@inproceedings{10.1145/3701716.3715245,
author = {Ren, Jiyuan and Du, Zhaocheng and Wen, Zhihao and Jia, Qinglin and Dai, Sunhao and Wu, Chuhan and Dong, Zhenhua},
title = {Few-shot LLM Synthetic Data with Distribution Matching},
year = {2025},
isbn = {9798400713316},
publisher = {Association for Computing Machinery},
address = {New York, NY, USA},
url = {https://doi.org/10.1145/3701716.3715245},
doi = {10.1145/3701716.3715245},
booktitle = {Companion Proceedings of the ACM on Web Conference 2025},
pages = {432–441},
numpages = {10},
location = {Sydney NSW, Australia},
series = {WWW '25}
}

@misc{havrilla2024surveyingeffectsqualitydiversity,
      title={Surveying the Effects of Quality, Diversity, and Complexity in Synthetic Data From Large Language Models}, 
      author={Alex Havrilla and Andrew Dai and Laura O'Mahony and Koen Oostermeijer and Vera Zisler and Alon Albalak and Fabrizio Milo and Sharath Chandra Raparthy and Kanishk Gandhi and Baber Abbasi and Duy Phung and Maia Iyer and Dakota Mahan and Chase Blagden and Srishti Gureja and Mohammed Hamdy and Wen-Ding Li and Giovanni Paolini and Pawan Sasanka Ammanamanchi and Elliot Meyerson},
      year={2024},
      eprint={2412.02980},
      archivePrefix={arXiv},
      primaryClass={cs.LG},
      url={https://arxiv.org/abs/2412.02980}, 
}

@inproceedings{10.5555/3666122.3667604,
author = {Xie, Sang Michael and Santurkar, Shibani and Ma, Tengyu and Liang, Percy},
title = {Data selection for language models via importance resampling},
year = {2023},
publisher = {Curran Associates Inc.},
address = {Red Hook, NY, USA},
booktitle = {Proceedings of the 37th International Conference on Neural Information Processing Systems},
articleno = {1482},
numpages = {27},
location = {New Orleans, LA, USA},
series = {NIPS '23}
}

@inproceedings{10.5555/3737916.3738886,
author = {Penedo, Guilherme and Kydl\'{\i}\v{c}ek, Hynek and Allal, Loubna Ben and Lozhkov, Anton and Mitchell, Margaret and Raffel, Colin and Von Werra, Leandro and Wolf, Thomas},
title = {The FineWeb datasets: decanting the web for the finest text data at scale},
year = {2024},
isbn = {9798331314385},
publisher = {Curran Associates Inc.},
address = {Red Hook, NY, USA},
booktitle = {Proceedings of the 38th International Conference on Neural Information Processing Systems},
articleno = {970},
numpages = {39},
location = {Vancouver, BC, Canada},
series = {NIPS '24}
}

@article{DBLP:journals/corr/abs-2512-02556,
  author       = {DeepSeek{-}AI},
  title        = {DeepSeek-V3.2: Pushing the Frontier of Open Large Language Models},
  journal      = {CoRR},
  volume       = {abs/2512.02556},
  year         = {2025},
  url          = {https://doi.org/10.48550/arXiv.2512.02556},
  doi          = {10.48550/ARXIV.2512.02556},
  eprinttype   = {arXiv},
  eprint       = {2512.02556},
  bibsource    = {dblp computer science bibliography, https://dblp.org}
}

@article{DBLP:journals/corr/abs-2604-15804,
  author       = {Qwen{-}Team},
  title        = {Qwen3.5-Omni Technical Report},
  journal      = {CoRR},
  volume       = {abs/2604.15804},
  year         = {2026},
  url          = {https://doi.org/10.48550/arXiv.2604.15804},
  doi          = {10.48550/ARXIV.2604.15804},
  eprinttype   = {arXiv},
  eprint       = {2604.15804},
  bibsource    = {dblp computer science bibliography, https://dblp.org}
}

@inproceedings{reimers2019sentencebertsentenceembeddingsusing,
    title = "Sentence-{BERT}: Sentence Embeddings using {S}iamese {BERT}-Networks",
    author = "Reimers, Nils  and
      Gurevych, Iryna",
    editor = "Inui, Kentaro  and
      Jiang, Jing  and
      Ng, Vincent  and
      Wan, Xiaojun",
    booktitle = "Proceedings of the 2019 Conference on Empirical Methods in Natural Language Processing and the 9th International Joint Conference on Natural Language Processing (EMNLP-IJCNLP)",
    month = nov,
    year = "2019",
    address = "Hong Kong, China",
    publisher = "Association for Computational Linguistics",
    url = "https://aclanthology.org/D19-1410/",
    doi = "10.18653/v1/D19-1410",
    pages = "3982--3992"
}

@misc{aminabadi2022deepspeedinferenceenablingefficient,
      title={DeepSpeed Inference: Enabling Efficient Inference of Transformer Models at Unprecedented Scale}, 
      author={Reza Yazdani Aminabadi and Samyam Rajbhandari and Minjia Zhang and Ammar Ahmad Awan and Cheng Li and Du Li and Elton Zheng and Jeff Rasley and Shaden Smith and Olatunji Ruwase and Yuxiong He},
      year={2022},
      eprint={2207.00032},
      archivePrefix={arXiv},
      primaryClass={cs.LG},
      url={https://arxiv.org/abs/2207.00032}, 
}

@misc{zheng2024llamafactoryunifiedefficientfinetuning,
      title={LlamaFactory: Unified Efficient Fine-Tuning of 100+ Language Models}, 
      author={Yaowei Zheng and Richong Zhang and Junhao Zhang and Yanhan Ye and Zheyan Luo and Zhangchi Feng and Yongqiang Ma},
      year={2024},
      eprint={2403.13372},
      archivePrefix={arXiv},
      primaryClass={cs.CL},
      url={https://arxiv.org/abs/2403.13372}, 
}

@article{DBLP:journals/corr/abs-2505-09388,
  author       = {Qwen{-}Team},
  title        = {Qwen3 Technical Report},
  journal      = {CoRR},
  volume       = {abs/2505.09388},
  year         = {2025},
  url          = {https://doi.org/10.48550/arXiv.2505.09388},
  doi          = {10.48550/ARXIV.2505.09388},
  eprinttype   = {arXiv},
  eprint       = {2505.09388},
  bibsource    = {dblp computer science bibliography, https://dblp.org}
}

@article{team2025every,
  title={Every activation boosted: Scaling general reasoner to 1 trillion open language foundation},
  author={Team, Ling and Li, Ang and Liu, Ben and Hu, Binbin and Li, Bing and Zeng, Bingwei and Ye, Borui and Tang, Caizhi and Tian, Changxin and Huang, Chao and others},
  journal={arXiv preprint arXiv:2510.22115},
  year={2025}
}

@article{grattafiori2024llama,
  title={The llama 3 herd of models},
  author={Grattafiori, Aaron and Dubey, Abhimanyu and Jauhri, Abhinav and Pandey, Abhinav and Kadian, Abhishek and Al-Dahle, Ahmad and Letman, Aiesha and Mathur, Akhil and Schelten, Alan and Vaughan, Alex and others},
  journal={arXiv preprint arXiv:2407.21783},
  year={2024}
}

\newpage

\appendix

\section{Dataset Statistics}
\label{Dataset Statistics}


We report the statistics of the six knowledge evaluation benchmarks used in this work in Tab.~\ref{tab:dataset_statistics}, including Web Questions, Natural Questions, TriviaQA, SimpleQA, SimpleQA-Verified, and EntityQuestions. The ``Candidate Answer'' column indicates whether the dataset provides multiple candidate answers.

\begin{table}[h]
\centering
\resizebox{1.0\columnwidth}{!}{%
\begin{tabular}{lcc}
\hline
Dataset & Number of Test Set & Candidate Answer \\ \hline
Web Questions & 2032 & yes \\
Natural Questions & 3610 & yes \\
TriviaQA & 8837 & yes \\
SimpleQA & 4326 & no \\
SimpleQA-Verified & 1000 & no \\
EntityQuestions & 12452 & yes \\ \hline
\end{tabular}
}
\caption{Dataset Statistics of Web Questions, Natural Questions, TriviaQA, SimpleQA, SimpleQA-Verified, EntityQuestions.} 
\label{tab:dataset_statistics}
\end{table}

\section{Prompt Details}
\label{Prompt Details}

\subsection{Knowledge Points Extraction Prompt}
\label{Knowledge Points Extraction Prompt}

\begin{Promptbox}{Knowledge Points Extraction Prompt}

You are an expert in logical reasoning and knowledge structure analysis. 

Given a Question and its Answer, your task is to: 

1. **Extract the Topic Entity/Entities** — the core subject(s) or object(s) the question is about. There may be one or multiple topic entities.

2. **List All Relevant Knowledge Points** — concisely list all knowledge points, facts, concepts, or information helpful for answering the question.

3. **Abstract the Logic Form Between Knowledge Points** — represent how the knowledge points are connected through the reasoning structure or logical flow used to arrive at the answer.

4. **Output the Logic Form in Mermaid format** — choose the most appropriate diagram type from the following: 

- **flowchart** (`flowchart TD` or `flowchart LR`) — for sequential reasoning, decision-making processes, or step-by-step logic 

- **graph** (`graph TD` or `graph LR`) — for relationship networks, entity connections, or multi-directional reasoning 

- **mindmap** (`mindmap`) — for hierarchical concept breakdowns, category exploration, or radial thinking patterns 

5. **Enable comparison** — your abstraction should allow detection of overlap in Logic Forms across different questions. 

**Selection Guidelines:** 

- Use **flowchart** when reasoning follows a clear sequence or involves conditional branches 

- Use **graph** when showing interconnected relationships or bidirectional reasoning paths 

- Use **mindmap** when the logic expands from a central concept into subcategories or attributes 

--- Analyze the Logic Form from the Following --- 

Question: \$question

Answer: \$answer

**Output Format (JSON):**
```json
{

`topic\_entities': [`Entity1', `Entity2', `...'],

"knowledge\_points": ["Knowledge point 1", "Knowledge point 2", "..."],

"logic\_form": "

```mermaid

[Mermaid diagram code]

```"

}
\end{Promptbox}

\subsection{Synthetic Prompt}
\label{Synthetic Prompt}

\begin{Promptbox}{Synthetic Prompt}

\# System:

You are an expert in educational content synthesis and knowledge representation. Your task is to create novel, logically coherent questions by intelligently combining elements from multiple input questions through deep analysis of underlying knowledge structures. **All synthesized questions MUST be instance-based, examining specific concrete entities (e.g., specific people, events, places, organizations, cases) rather than abstract concepts or vague generalities.**

\#\# Core Design Principles:

0. **Instance-Based Questions (MANDATORY)**:

- ALL synthesized questions MUST examine specific, concrete instances

- Questions MUST involve particular entities: specific people, events, places, organizations, dates, cases, or other concrete subjects

- PROHIBIT abstract conceptual questions or vague generalities

- Example ACCEPTABLE: "In which year did Marie Curie win her first Nobel Prize?"

- Example PROHIBITED: "What are the characteristics of scientific achievement?"

1. **Question Differentiation**:

- Generated questions MUST differ substantially from source questions in content and presentation

- Knowledge points in new questions MUST derive from source knowledge points or their thematic domains

- Avoid superficial variations; ensure genuine conceptual recombination

2. **Question Diversity**:

- Explore different aspects, angles, and dimensions of the knowledge domain

- Prevent structural or thematic homogeneity across generated questions

- All questions MUST use Q\&A format WITHOUT multiple-choice options

3. **Logical Consistency**:

- Ensure coherence among question, knowledge points, and answer

- **Eliminate mechanical combinations and logical contradictions**

4. **Distractor Integration**:

- Include distractor elements derived from source knowledge points or thematic domains

- Design distractors to increase question difficulty meaningfully

5. **Quality Validation** (MANDATORY for each question):

Conduct three critical checks before detailed evaluation. Apply STRICT standards for all checks - only flag obvious violations. When uncertain, allow the question to proceed to full evaluation.

**Preliminary Checks:**

**A. Answer Independence (answer\_directly\_in\_question)**

- Is the answer explicitly stated in the question text?

- FAIL (false): "The Jammu and Kashmir State Film Development Corporation is focused on promoting cinema in which Indian union territory?" → "Jammu and Kashmir"

- PASS (true): "What is the capital of France?" → "Paris"

**B. Answer Verifiability (answer\_verifiable)**

- Can the answer be objectively verified?

- FAIL (false): "What is the most beautiful color?" → "Blue"

- PASS (true): "What is the boiling point of water at sea level?" → "100°C"

**C. Answer Correctness (answer\_correctness)**

- Based on your knowledge, is the answer correct?

- Return "false" ONLY when absolutely certain the answer is incorrect

- Return "true" when the answer is correct

- Return "unknown" when you cannot verify with high confidence

- FAIL (false): "What is the capital of France?" → "London"

- PASS (true): "What is 2 + 2?" → "4"

- UNCERTAIN (unknown): "What was the population of San Diego in 1987?" → "Approximately 2.24 million"

**CRITICAL RULE: If either answer\_directly\_in\_question or answer\_verifiable fails, OR if answer\_correctness is false, IMMEDIATELY DISCARD this question and REGENERATE a new question.** If ALL preliminary checks pass (including answer\_correctness being true or unknown), proceed with the full evaluation below.

**Detailed Evaluation Criteria:**

**5.1. Educational Significance**

- Does the question contain meaningful, valuable knowledge worth learning?

- Is the difficulty level appropriate for educational purposes?

- Does it promote critical thinking or understanding of important concepts?

**5.2. Specificity and Concreteness**

- Does the question examine specific, concrete knowledge rather than only abstract concepts?

- Does it involve particular instances such as specific people, events, places, or cases?

**5.3. Internal Question Logic**

- Is the question itself logically coherent and well-structured?

- Are the premises, conditions, and requirements clearly stated?

**5.4. Question-Answer Logic**

- Does the answer logically follow from the question?

- Is the reasoning path from question to answer valid and complete?

**5.5. Knowledge-Point Relevance \& Logic Diagram Completeness**

- Are the knowledge points relevant and sufficient for answering the question?

- Does the logic diagram form a complete reasoning chain?

- Are knowledge points and diagram steps consistent with each other?

\#\# Synthesis Process:

**Step 1: Knowledge Point Extraction**

For each input question, comprehensively identify 10-20 knowledge points including:

- Direct knowledge required for answering (MUST be specific instance-based knowledge points, e.g., "Marie Curie won Nobel Prize in Physics in 1903" NOT "characteristics of Nobel Prize winners")

- Related domain/topic information (MUST reference specific entities, events, or cases)

- Contextual and background knowledge (MUST involve concrete instances)

- **MANDATORY**: ALL extracted and extended knowledge points MUST be instance-based, referring to specific entities, events, people, places, or concrete cases

- **PROHIBIT**: Abstract conceptual knowledge points or vague generalities

This forms the source pool for synthesis.

**Step 2: Knowledge Point Combination \& Validation**

- Attempt cross-question knowledge point combinations to synthesize new questions

- **CRITICAL**: Validate EACH synthesized question against Quality Validation criteria

- If validation fails, IMMEDIATELY DISCARD and REGENERATE the question

- **If REGENERATED question still fails validation after multiple attempts, ABANDON synthesis of this particular question and proceed with remaining questions**

- If cross-question combination is difficult, synthesize from single-question knowledge points

- Ensure questions derive from diverse knowledge points and perspectives

- **IMPORTANT**:

- The core knowledge points required for synthesized questions MUST avoid being identical to those of the original questions, but should revolve around the same thematic domain

- ALL synthesized questions MUST be instance-based, examining specific concrete entities (people, events, places, organizations, cases)

- **PROHIBIT**:

- Counterfactual assumptions or open-ended subjective questions

- Abstract conceptual questions or vague generalities (e.g., "What are the main features of...", "What is the general principle of...")

- Forced, artificial combinations lacking educational value

- Questions that mechanically merge unrelated domains without meaningful conceptual connection (e.g., "What is the relationship between the targeted delivery mechanism of Antibody-Drug Conjugates and the organizational structure of the Oceania Football Confederation in terms of specialized components working toward specific objectives?")

**Step 3: JSON Output**

Output synthesized questions in the specified JSON format. **Target 5 questions; fewer is acceptable if quality standards cannot be met.**

\#\# Output Format Requirements:

Return a JSON array with synthesized question objects (target 5; fewer if necessary to maintain quality), each containing:

- `"synthesized\_question"`: The newly created question (MUST be instance-based)

- `"answer"`: The correct answer

- `"topic\_entities"`: Array of main entities

- `"knowledge\_points"`: Array of 3-5 key knowledge points needed to answer (MUST be instance-based)

- `"knowledge\_logic"`: Complete Mermaid diagram (as string) showing reasoning path

\# User:

I will provide k=3 questions in JSON format. Synthesize new questions by intelligently combining their elements. Follow system instructions precisely.

\#\# Input Questions:

\$input\_question\_list

Synthesize questions (target 5; fewer if necessary to maintain quality) ensuring each has:

1. Clear, answerable question with correct answer (MUST be instance-based, examining specific entities)

2. Relevant topic entities

3. 3-5 knowledge points (MUST be instance-based)

4. Complete Mermaid logic diagram

Return your response including:

- **Step 1**: Knowledge Point Expansion (ALL knowledge points MUST be instance-based)

- **Step 2**: Knowledge Point Combination \& Validation (with regeneration if validation fails; abandon specific question synthesis if regeneration repeatedly fails)

- **Final Output**: JSON array with synthesized question objects (target 5; fewer if necessary to maintain quality) in specified format

\end{Promptbox}

\subsection{Evaluation Prompt}
\label{Evaluation Prompt}

Synthesized candidate questions may exhibit various quality issues, such as ambiguous intent, meaningless content, hallucinated answers, incorrect knowledge point lists, or formatting errors. We design a detailed evaluation prompt to filter out low-quality samples that do not meet our requirements. Below we describe the rationale and role of each evaluation dimension. We use DeepSeek V3.2 and Qwen3.5-397B-A17B as LLM judges, following a two-step evaluation pipeline.

\paragraph{Preliminary Check.} Before scoring, each sample undergoes a preliminary check across three binary criteria. \textbf{Answer Independence} verifies that the answer is not directly inferable from the question itself, ensuring the question genuinely tests knowledge. \textbf{Answer Verifiability} ensures the answer is objective and verifiable, excluding questions with subjective or opinion-based answers. \textbf{Answer Correctness} filters out samples containing factual errors or common-sense mistakes. Any sample that fails on any of these three criteria is immediately discarded without further evaluation.

\paragraph{Scoring.} Samples that pass the preliminary check are then scored across five dimensions (total score: 12). \textbf{Educational Significance} (0\textasciitilde4) measures whether the question contains meaningful knowledge worth learning, penalizing trivial or content-free questions. \textbf{Specificity and Concreteness} (0\textasciitilde2) assesses whether the question targets specific, concrete knowledge rather than overly abstract concepts, encouraging instance-level questions. \textbf{Internal Question Logic} (0\textasciitilde2) checks whether the question itself is logically coherent and well-structured, as combining multiple knowledge points may sometimes result in forced or incoherent compositions. \textbf{Question-Answer Logic} (0\textasciitilde2) evaluates whether the answer logically follows from the question, ensuring the reasoning chain between question and answer is sound. \textbf{Knowledge-Point Relevance \& Logic Diagram Completeness} (0\textasciitilde2) jointly checks whether the associated knowledge points are relevant and sufficient for answering the question and whether the logic diagram forms a complete reasoning chain.

Samples with a score of 0 on any single dimension are excluded, and only samples with a final score $\geq 8$—averaged across the two LLM judges—are retained. The LLM judge prompt is provided below.

\begin{Promptbox}{Evaluation Prompt}

System:

You are a specialized educational question evaluator. Your task is to assess the quality and logical structure of educational questions by analyzing the question itself, its answer, knowledge points, and logic diagram. Focus on verifiability, educational value, and logical coherence.

User:

I will provide you with a question evaluation request containing four elements:

1. The question text

2. The answer

3. The knowledge points (array of strings)

4. The logic form (Mermaid diagram as string)

**PRELIMINARY CHECK:**

Conduct three critical checks before detailed evaluation. Apply STRICT standards for all checks - only flag obvious violations. When uncertain, allow the question to proceed to full evaluation.

**1. Answer Independence (answer\_directly\_in\_question)**

Is the answer directly contained in the question text? (**Return fail ONLY if the exact answer text appears literally within the question text**)

- FAIL example (fail):

Question: "The Jammu and Kashmir State Film Development Corporation is focused on promoting cinema in which Indian union territory?"

Answer: "Jammu and Kashmir" (The answer 'Jammu and Kashmir' appears directly at the start of the question in 'The Jammu and Kashmir State Film Development Corporation')

- PASS example (pass):

Question: "What is the capital of France?"

Answer: "Paris"

**2. Answer Verifiability (answer\_verifiable)**

Can the answer be objectively verified? (Return fail only when the answer is a completely subjective response rather than an objective factual response)

- FAIL example (fail):

Question: "What is the most beautiful color?"

Answer: "Blue"

- PASS example (pass):

Question: "What is the boiling point of water at sea level?"

Answer: "100°C (212°F)"

**3. Answer Correctness (answer\_correctness)**

Based on your knowledge, is the answer correct?

- Return "fail" ONLY when you are absolutely certain the answer is incorrect

- Return "pass" when the answer is correct

- Return "unknown" when you cannot verify with high confidence

Examples:

- FAIL (fail):

Question: "What is the capital of France?"

Answer: "London"

- PASS (pass):

Question: "What is 2 + 2?"

Answer: "4"

- UNCERTAIN (unknown):

Question: "What was the population of a small town in 1987?"

Answer: "12,453"

**CRITICAL RULE: If either answer\_directly\_in\_question or answer\_verifiable fails (returns fail), OR if answer\_correctness is fail, set all scores to 0 and skip detailed evaluation.** If ALL preliminary checks pass (including answer\_correctness being pass or unknown), proceed with the full evaluation below.

**Evaluation Criteria:**

**1. Educational Significance (0-4 points)**

- Does the question contain meaningful, valuable knowledge worth learning?

- Is the difficulty level appropriate for educational purposes?
- Does it promote critical thinking or understanding of important concepts?

Examples:

- Score 0: No educational value or trivial content

Example: "Considering the typical structure of a scientific article, in which section would you most likely find the detailed experimental protocol used to verify the principles underlying Listing's Law?"
Answer: "Methods"

- Score 1: Minimal educational value, very basic or irrelevant knowledge

Example: "A scholar is conducting primary source research for a dissertation on the military strategies used during the Bangladesh Liberation War. At which type of higher education institution is this scholar most likely employed?"

Answer: "A research university"

- Score 2: Some educational value but limited depth or applicability
Example: "What is the difference in years between the publication of a book by co-authors and the year a Mughal prince was sentenced, if the book was published in 1980 and the sentencing year was 1661?"

Answer: "319"

- Score 3: Good educational value with meaningful knowledge

Example: "In what year was the statue 'Amazone' completed?"

Answer: "1923"

- Score 4: Excellent educational value, promotes deep understanding

Example: "In the context of American industry and arts, which state serves as the primary historical hub for both large-scale automobile manufacturing and the foundational recording studio for a famous soul ballad singer like Anita Baker?"

Answer: "Michigan"

**2. Specificity and Concreteness (0-2 points)**

- Does the question examine specific, concrete knowledge rather than only abstract concepts?

- Does it involve particular instances such as specific people, events, places, or cases?

Examples:

- Score 0: Question is purely conceptual/abstract with no concrete examples or specific instances

Example: "What type of urban area is most likely to house both a city's main public library and a significant equestrian statue commemorating a national leader?" (purely conceptual, overly broad)

Example: "If an artist recognized for contributions to cultural diplomacy received a major U.S. State Department award in the same decade that a key lunar precursor program ended, in what year did that artist likely receive the award?" (purely conceptual, overly broad)

- Score 1: Question includes some specific elements but remains largely conceptual or vague

Example: "Drawing an analogy to a structured legal code created for societal benefit, what is the primary purpose of a comprehensive, annotated bibliography like 'British Literary Bibliographies'?" (conceptual with some structure)

Example: "For a historically significant film released in 2006 about a 1947 event, which renowned composer, known for setting records in his field, would be a plausible candidate to score its music, given his profile of working on major studio productions?" (vague, lacks specific description)

- Score 2: Question clearly examines concrete, specific knowledge (e.g., specific people, particular historical figures, specific events, named locations, real cases)

Example: "What was the specific honor received by Muhammad Ahmad Said Khan Chhatari in the year 1946 that represented his final recognition in the British honors system?" (specific person/event)

Example: "Beyond their role in affective cognition, the anterior thalamic nuclei are a critical component of a well-known neural circuit associated with memory. What is the name of this circuit?" (specific case)

**3. Internal Question Logic (0-2 points)**

- Is the question itself logically coherent and well-structured?

- Are the premises, conditions, and requirements clearly stated?

Examples:

- Score 0: Question is illogical, contradictory, or unclear
Example: "For a database system tracking legal cases like that of a football executive awaiting trial, which capability of a hardware accelerator would be most critical for generating timely reports on case status?" (forced analogy, logically incoherent)

Example: "If the section 'Finding the Angle When the Function Is Given' were published in a year that is the sum of the maximum stowable people and the number of floors in the west wing, what year would it be?" (knowledge points span too broadly, their connection lacks logic)

- Score 1: Question has minor logical issues or ambiguities

Example: "The individual who served three times as MP for Rochester first entered Parliament in the same decade that significant updates were made to which game's rules?"

Example: "The end of BSA's dual-city car production in 1939 and the 18th-century origin of the 'Mediterranean paradise' concept both relate to significant transitions in their respective fields. What major global event beginning in 1939 likely precipitated the end of this automotive manufacturing phase?"

- Score 2: Question is logically sound and clearly structured

Example: "Following his family's relocation to Lademoen, in which Norwegian city did Hjalmar Andersen develop his speed skating career that led to his Olympic fame?"

Example: "Both the Munch Museum and the neighborhood of Lademoen are located in major Norwegian cities. One is in the capital, and the other is in a historical city known for its university and as a former capital. What is the primary geographical distinction between their host cities?"

**4. Question-Answer Logic (0-2 points)**

- Does the answer logically follow from the question?

- Is the reasoning path from question to answer valid and complete?

- Score 0: Answer doesn't logically follow from question or contains fallacies

- Score 1: Answer mostly follows but has logical gaps or weaknesses

- Score 2: Answer logically and directly follows from the question

**5. Knowledge-Point Relevance \& Logic Diagram Completeness (0-2 points)**

- Are the knowledge points relevant and sufficient for answering the question?

- Does the logic diagram form a complete reasoning chain?

- Are knowledge points and diagram steps consistent with each other?

- Score 0: Knowledge points irrelevant or diagram incomplete/disconnected

- Score 1: Most elements present but with gaps or inconsistencies

- Score 2: Knowledge points fully relevant and diagram provides complete logical path

**Scoring Summary:**

- 0-3: Poor quality (fundamental issues, not suitable for educational use)

- 4-6: Fair quality (significant issues but salvageable)

- 7-9: Good quality (minor improvements needed)

- 10-12: Excellent quality (high educational and logical value)

**Output Format:**

Provide your evaluation as a JSON object with the following structure:

{

"total\_score": [integer 0-12],

"criterion\_scores": {

"educational\_significance": [integer 0-4],

"specificity\_and\_concreteness": [integer 0-2],

"internal\_question\_logic": [integer 0-2],

"question\_answer\_logic": [integer 0-2],

"knowledge\_and\_diagram\_quality": [integer 0-2]

},

"justifications": {

"educational\_significance": "[brief justification]",

"specificity\_and\_concreteness": "[brief justification]",

"internal\_question\_logic": "[brief justification]",

"question\_answer\_logic": "[brief justification]",

"knowledge\_and\_diagram\_quality": "[brief justification]"

},

"preliminary\_checks": {

"answer\_directly\_in\_question": ["pass" | "fail"],

"answer\_verifiable": ["pass" | "fail"],

"answer\_correctness": ["pass" | "fail" | "unknown"],

"preliminary\_check": ["pass" | "fail"]

},

"overall\_assessment": "[Poor/Fair/Good/Excellent]",

"main\_strength": "[single strongest aspect or 'N/A' if failed preliminary]",

"main\_weakness": "[single most important improvement needed]"

}

**Input to Evaluate:**

{

"question": \$question,

"answer": \$answer,

"knowledge\_points": \$knowledge\_points,

"logic\_form": \$knowledge\_logic

}

Evaluate this input strictly according to the preliminary checks and the 5 criteria above. If any preliminary check fails, set all scores to 0 and indicate the failure reason. Provide only the JSON output.

\end{Promptbox}

\section{Supplement Implementation Details}
\label{Supplement Implementation Details}

\paragraph{Details of data synthesis.}
We use DeepSeek V3.2 for knowledge point extraction and data synthesis, and use both DeepSeek V3.2 and Qwen3.5-397B-A17B as LLM judges for quality filtering, with the average score of the two used for sample selection. We collect approximately 14M seed QA pairs (1.73B tokens) from Wikipedia, which KDoS expands to 71M samples (9.28B tokens, approximately 125 tokens per sample), with a maximum iteration count of $k=200$ and convergence thresholds $\epsilon^{\rho}$ and $\epsilon^{T}$ both set to 1\%. 
\paragraph{Details of experiment settings.}
Data synthesis and quality filtering are conducted on an NVIDIA H20-3E cluster with 128 nodes $\times$ 8 GPUs (1024 H20-3E GPUs in total), achieving a synthesis throughput of 2,075 instances per 8-GPU node per hour and a quality filtering throughput of 15,700 instances per 8-GPU node per hour. Knowledge injection experiments are conducted on models including Qwen3.0-base, Ling-mini-2.0-base, and LLaMA-3.2-base, with Qwen3-4B-Base as the default backbone, using an NVIDIA H800 cluster with 8 nodes $\times$ 8 GPUs (64 H800 GPUs in total). We use LLaMA-Factory \citep{zheng2024llamafactoryunifiedefficientfinetuning} as the training framework. For Qwen3.0-base and LLaMA-3.2-base, we use DeepSpeed \citep{aminabadi2022deepspeedinferenceenablingefficient} ZeRO-3 for acceleration; for Ling-mini-2.0-base, due to its Mixture-of-Experts (MoE) architecture, we use ZeRO-2 instead. The learning rate is set to $1\times10^{-5}$ with a cosine decay schedule, and each model is trained for 1 epoch over the full dataset. The density range used in our experiments, $10^{-4}$\textasciitilde$10^{36}$, depends on the choice of embedding model. Specifically, we use \texttt{sentence-transformers/all-MiniLM-L6-v2} to embed knowledge points into an $n=384$-dimensional space. Since the density formula in Eq.~\ref{eq:density} is sensitive to $n$, we adopt this range to ensure full coverage of our 71M synthesis pool and clear separation across different distributions. We emphasize that $\rho$ is not an artifact of the embedding model: it measures the token-to-volume ratio in semantic space, and the underlying data distribution is model-agnostic. Different embedding models would change the absolute numeric scale of $\rho$ due to different $n$, but the relative distributional structure of the data and the existence of an optimal density range remain invariant. The specific values $10^{-4}$\textasciitilde$10^{4}$ reported as optimal are tied to our embedding choice; the general principle that an optimal knowledge density range exists holds regardless.
\paragraph{Details of Baselines.}
\textbf{Rand.} synthesizes questions directly from the seed pool without any selection criterion and stops once the target token count $T^{\text{target}}$ is reached.
\textbf{Uni.} clusters all candidate samples into domains via K-Means in semantic space and enforces an equal token quota across all domains, ensuring a uniform token distribution over knowledge domains.
\textbf{Diff.} follows the same synthesis process as Rand., but applies difficulty-weighted importance selection. In each iteration, a batch of candidate samples is synthesized and scored by the base model $\mathcal{M}$ via perplexity (PPL). The top 60\% by higher PPL are directly accepted; the remaining samples are added to a candidate pool. In subsequent iterations, 60\% of accepted samples are drawn from the current batch and the rest from the candidate pool, both ranked by PPL. This continues until $T^{\text{target}}$ is reached.
\textbf{Qual.} also follows the same synthesis process as Rand., but applies quality-based rejection selection. In each iteration, a batch of candidates is scored by LLM judges. The top 60\% by quality score are directly accepted; the remaining samples are added to a candidate pool. In subsequent iterations, 60\% of accepted samples are drawn from the current batch and the rest from the candidate pool, both ranked by quality score. This continues until $T^{\text{target}}$ is reached.

\setlength{\algomargin}{6pt}
\SetInd{0em}{1.2em}
\linespread{1.2}
\begin{algorithm}[h]
\small
\SetKw{continue}{continue}
\SetKwInOut{Input}{Input}
\SetKwInOut{Output}{Output}

\Input{Quality-filtered candidate pool $\mathcal{S}^{\text{pass}}$, current data pool $\mathcal{S}^{\text{syn}}$, target token count $T^{\text{target}}$, target density $\rho^{\text{target}}$, max iterations $k$, convergence thresholds $\epsilon^{T}$, $\epsilon^{\rho}$}
\Output{Synthetic data pool $\mathcal{S}^{\text{syn}} = (T^{\text{target}}, \rho^{\text{target}})$}

\# Pre-compute $r^{\text{target}}$ from $T^{\text{target}}$ and $\rho^{\text{target}}$ via Eq.~\ref{eq:density}

$r^{\text{target}} \leftarrow \left( \dfrac{T^{\text{target}} \cdot \Gamma(n/2+1)}{\pi^{n/2} \cdot \rho^{\text{target}}} \right)^{1/n}$;

$t \leftarrow 0$;

\While{$t < k$ \textbf{ and } $( |T - T^{\text{target}}| \geq \epsilon^{T}$ \textbf{ or } $|\rho - \rho^{\text{target}}| \geq \epsilon^{\rho} )$}{

    Compute current $T \leftarrow |\mathcal{S}^{\text{syn}}|_{\text{tokens}}$\;
    
    Compute current $r \leftarrow$ mean distance of samples in $\mathcal{S}^{\text{syn}}$ to centroid\;

    \ForEach{candidate $c \in \mathcal{S}^{\text{pass}}$}{

        \eIf{$T < T^{\text{target}}$}{

            \# \textbf{Cold-start phase:} accumulate data volume without density constraint

            $\mathcal{S}^{\text{syn}} \leftarrow \mathcal{S}^{\text{syn}} \cup \{c\}$\;

        }{

            \# \textbf{Density fine-tuning phase:} accept based on $r$ vs.\ $r^{\text{target}}$

            $d_c \leftarrow$ distance from $c$ to centroid of $\mathcal{S}^{\text{syn}}$\;

            \eIf{$r < r^{\text{target}}$}{

                \# Density too high: prefer samples far from centroid to increase $r$

                Accept $c$ with probability $\propto d_c$\;

            }{

                \# Density too low: prefer samples close to centroid to decrease $r$

                Accept $c$ with probability $\propto 1 / d_c$\;

            }

            \If{$c$ accepted}{
                $\mathcal{S}^{\text{syn}} \leftarrow \mathcal{S}^{\text{syn}} \cup \{c\}$\;
            }
        }

        Update $T$, $\rho$, $r$ of $\mathcal{S}^{\text{syn}}$\;
    }

    $t \leftarrow t + 1$\;
}

\Return{$\mathcal{S}^{\text{syn}}$};

\caption{Rejection Sampling in KDoS}
\label{algorithm:rejection_sampling}
\end{algorithm}

\section{Supplement Experimental Evaluation Details}
\label{Supplement Experimental Evaluation Details}

Below we provide the detailed numerical results and complete visualizations for each group of experiments.

\subsection{Supplement Experimental Evaluation Loss Details}
\label{Supplement Experimental Evaluation Loss Details}

Below we provide the detailed eval loss values for each experiment in Sec.~\ref{scaling with model and data size}. Details are provided in Tab.~\ref{tab:model_size_1b_qwen3_0.6b}, \ref{tab:model_size_1b_qwen3_1.7b}, \ref{tab:model_size_1b_qwen3_4b}, \ref{tab:model_size_1b_qwen3_8b}, \ref{tab:model_size_1b_qwen3_14b}, \ref{tab:data_size_3b_qwen3_4b}, and \ref{tab:data_size_5b_qwen3_4b}.

\begin{table*}[ht]
\centering
\resizebox{1.0\textwidth}{!}{%
\begin{tabular}{lccccccc}
\hline
Density & entity-questions & simpleqa & simpleqa-verified & webquestions & nq & triviaqa & total loss \\ \hline
1e-4 & 2.0414 & 1.7918 & 1.9259 & 2.3364 & 2.2244 & 3.2438 & 13.5637 \\
1 & 2.0611 & 1.7900 & 1.9356 & 2.3750 & 2.2401 & 3.2346 & 13.6364 \\
1e4 & 2.0634 & 1.7877 & 1.9331 & 2.3195 & 2.2107 & 3.2365 & 13.5509 \\
1e12 & 2.0827 & 1.7984 & 1.9505 & 2.3336 & 2.2031 & 3.2264 & 13.5947 \\
1e20 & 2.1459 & 1.8179 & 1.9684 & 2.4282 & 2.2515 & 3.2651 & 13.8770 \\
1e28 & 2.1708 & 1.8578 & 1.9985 & 2.4594 & 2.2856 & 3.3338 & 14.1059 \\
1e36 & 2.6982 & 2.3371 & 2.5184 & 3.1991 & 2.8715 & 3.9612 & 17.5859 \\ \hline
\end{tabular}
}
\caption{Eval loss of knowledge injection on Qwen3-0.6B-Base (1B training tokens) across Web Questions, Natural Questions, TriviaQA, SimpleQA, SimpleQA-Verified, and EntityQuestions.} 
\label{tab:model_size_1b_qwen3_0.6b}
\end{table*}

\begin{table*}[ht]
\centering
\resizebox{1.0\textwidth}{!}{%
\begin{tabular}{lccccccc}
\hline
Density & entity-questions & simpleqa & simpleqa-verified & webquestions & nq & triviaqa & total loss \\ \hline
1e-4 & 1.7444 & 1.6024 & 1.7517 & 1.9471 & 1.9073 & 2.8846 & 11.8374 \\
1 & 1.7510 & 1.5998 & 1.7477 & 1.9488 & 1.9091 & 2.8853 & 11.8416 \\
1e4 & 1.7542 & 1.6031 & 1.7569 & 1.9288 & 1.8889 & 2.8750 & 11.8069 \\
1e12 & 1.7758 & 1.6127 & 1.7732 & 1.9266 & 1.8924 & 2.8643 & 11.8451 \\
1e20 & 1.8305 & 1.6360 & 1.7966 & 1.9480 & 1.9201 & 2.9076 & 12.0388 \\
1e28 & 1.8797 & 1.6914 & 1.8525 & 2.0043 & 1.9721 & 3.0175 & 12.4176 \\
1e36 & 2.3234 & 2.0498 & 2.3604 & 2.6000 & 2.5044 & 3.6752 & 15.5132 \\ \hline
\end{tabular}
}
\caption{Eval loss of knowledge injection on Qwen3-1.7B-Base (1B training tokens) across Web Questions, Natural Questions, TriviaQA, SimpleQA, SimpleQA-Verified, and EntityQuestions.} 
\label{tab:model_size_1b_qwen3_1.7b}
\end{table*}

\begin{table*}[ht]
\centering
\resizebox{1.0\textwidth}{!}{%
\begin{tabular}{lccccccc}
\hline
Density & entity-questions & simpleqa & simpleqa-verified & webquestions & nq & triviaqa & total loss \\ \hline
1e-4 & 1.5830 & 1.4223 & 1.5742 & 1.7320 & 1.7465 & 2.6586 & 10.7165 \\
1 & 1.6003 & 1.4245 & 1.5836 & 1.7243 & 1.7468 & 2.6590 & 10.7385 \\
1e4 & 1.6087 & 1.4330 & 1.5791 & 1.7225 & 1.7349 & 2.6481 & 10.7262 \\
1e12 & 1.6138 & 1.4411 & 1.6000 & 1.7306 & 1.7317 & 2.6405 & 10.7577 \\
1e20 & 1.6740 & 1.4693 & 1.6130 & 1.7851 & 1.7723 & 2.6779 & 10.9914 \\
1e28 & 1.7553 & 1.5800 & 1.7281 & 1.9196 & 1.8870 & 2.8598 & 11.7299 \\
1e36 & 2.3090 & 2.0928 & 2.2662 & 2.6262 & 2.4930 & 3.5296 & 15.3167 \\ \hline
\end{tabular}
}
\caption{Eval loss of knowledge injection on Qwen3-4B-Base (1B training tokens) across Web Questions, Natural Questions, TriviaQA, SimpleQA, SimpleQA-Verified, and EntityQuestions.} 
\label{tab:model_size_1b_qwen3_4b}
\end{table*}

\begin{table*}[ht]
\centering
\resizebox{1.0\textwidth}{!}{%
\begin{tabular}{lccccccc}
\hline
Density & entity-questions & simpleqa & simpleqa-verified & webquestions & nq & triviaqa & total loss \\ \hline
1e-4 & 1.4431 & 1.3626 & 1.5063 & 1.6185 & 1.6339 & 2.4817 & 10.0461 \\
1 & 1.4543 & 1.3719 & 1.5149 & 1.6512 & 1.6597 & 2.4866 & 10.1386 \\
1e4 & 1.4661 & 1.3770 & 1.5201 & 1.6338 & 1.6322 & 2.4909 & 10.1201 \\
1e12 & 1.4848 & 1.3724 & 1.5340 & 1.6029 & 1.6138 & 2.4730 & 10.0809 \\
1e20 & 1.5018 & 1.3903 & 1.5412 & 1.6474 & 1.6401 & 2.5092 & 10.2300 \\
1e28 & 1.5866 & 1.5008 & 1.6626 & 1.8
183 & 1.7446 & 2.6857 & 10.9986 \\
1e36 & 2.1483 & 2.0641 & 2.2484 & 2.5960 & 2.3998 & 3.5211 & 14.9777 \\ \hline
\end{tabular}
}
\caption{Eval loss of knowledge injection on Qwen3-8B-Base (1B training tokens) across Web Questions, Natural Questions, TriviaQA, SimpleQA, SimpleQA-Verified, and EntityQuestions.} 
\label{tab:model_size_1b_qwen3_8b}
\end{table*}

\begin{table*}[ht]
\centering
\resizebox{1.0\textwidth}{!}{%
\begin{tabular}{lccccccc}
\hline
Density & entity-questions & simpleqa & simpleqa-verified & webquestions & nq & triviaqa & total loss \\ \hline
1e-4 & 1.3513 & 1.2646 & 1.4374 & 1.5477 & 1.5357 & 2.3828 & 9.5195 \\
1 & 1.3583 & 1.2637 & 1.4505 & 1.5666 & 1.5275 & 2.3818 & 9.5484 \\
1e4 & 1.3505 & 1.2651 & 1.4214 & 1.5523 & 1.5214 & 2.3708 & 9.4815 \\
1e12 & 1.3635 & 1.2796 & 1.4479 & 1.5490 & 1.5141 & 2.3650 & 9.5191 \\
1e20 & 1.3513 & 1.3088 & 1.4603 & 1.5944 & 1.5399 & 2.4023 & 9.6570 \\
1e28 & 1.5318 & 1.4856 & 1.6531 & 1.8088 & 1.7258 & 2.6651 & 10.8702 \\
1e36 & 2.1028 & 2.0886 & 2.2147 & 2.5668 & 2.3988 & 3.5138 & 14.8855 \\ \hline
\end{tabular}
}
\caption{Eval loss of knowledge injection on Qwen3-14B-Base (1B training tokens) across Web Questions, Natural Questions, TriviaQA, SimpleQA, SimpleQA-Verified, and EntityQuestions.} 
\label{tab:model_size_1b_qwen3_14b}
\end{table*}

\begin{table*}[ht]
\centering
\resizebox{1.0\textwidth}{!}{%
\begin{tabular}{lccccccc}
\hline
Density & entity-questions & simpleqa & simpleqa-verified & webquestions & nq & triviaqa & total loss \\ \hline
1e-4 & 1.5883 & 1.3732 & 1.5061 & 1.7098 & 1.8029 & 2.6993 & 10.6797 \\
1 & 1.5975 & 1.3723 & 1.5051 & 1.7287 & 1.7687 & 2.6966 & 10.6689 \\
1e4 & 1.6042 & 1.3750 & 1.5023 & 1.7192 & 1.7758 & 2.6859 & 10.6624 \\
1e12 & 1.6126 & 1.3814 & 1.5213 & 1.7401 & 1.7792 & 2.6754 & 10.7101 \\
1e20 & 1.6500 & 1.4005 & 1.5270 & 1.7343 & 1.7801 & 2.6647 & 10.7566 \\
1e28 & 1.6693 & 1.4425 & 1.5595 & 1.7739 & 1.7983 & 2.6648 & 10.9083 \\
1e36 & 2.2294 & 2.1333 & 2.1951 & 2.5390 & 2.5247 & 3.4122 & 15.0337 \\ \hline
\end{tabular}
}
\caption{Eval loss of knowledge injection on Qwen3-4B-Base (3B training tokens) across Web Questions, Natural Questions, TriviaQA, SimpleQA, SimpleQA-Verified, and EntityQuestions.} 
\label{tab:data_size_3b_qwen3_4b}
\end{table*}

\begin{table*}[ht]
\centering
\resizebox{1.0\textwidth}{!}{%
\begin{tabular}{lccccccc}
\hline
Density & entity-questions & simpleqa & simpleqa-verified & webquestions & nq & triviaqa & total loss \\ \hline
1 & 1.6129 & 1.3096 & 1.4442 & 1.7082 & 1.7958 & 2.7011 & 10.5718 \\
1e4 & 1.5996 & 1.3012 & 1.4351 & 1.7070 & 1.7738 & 2.6869 & 10.5036 \\
1e12 & 1.5988 & 1.3091 & 1.4544 & 1.6942 & 1.7752 & 2.6829 & 10.5146 \\
1e20 & 1.5937 & 1.3318 & 1.4848 & 1.7315 & 1.7803 & 2.6931 & 10.6152 \\
1e28 & 1.6043 & 1.3581 & 1.5062 & 1.7393 & 1.7841 & 2.6939 & 10.6859 \\
1e36 & 1.6143 & 1.3976 & 1.5341 & 1.7370 & 1.7885 & 2.6947 & 10.7662 \\
1e44 & 2.2516 & 1.9151 & 2.0735 & 2.5331 & 2.5012 & 3.5991 & 14.8736 \\ \hline
\end{tabular}
}
\caption{Eval loss of knowledge injection on Qwen3-4B-Base (5B training tokens) across Web Questions, Natural Questions, TriviaQA, SimpleQA, SimpleQA-Verified, and EntityQuestions.} 
\label{tab:data_size_5b_qwen3_4b}
\end{table*}

\subsection{Visualization of Scaling with Model Size}
\label{Visualization of Scaling with Model Size}

\begin{figure*}[ht]
    \centering
    \includegraphics[width=1.0\textwidth]{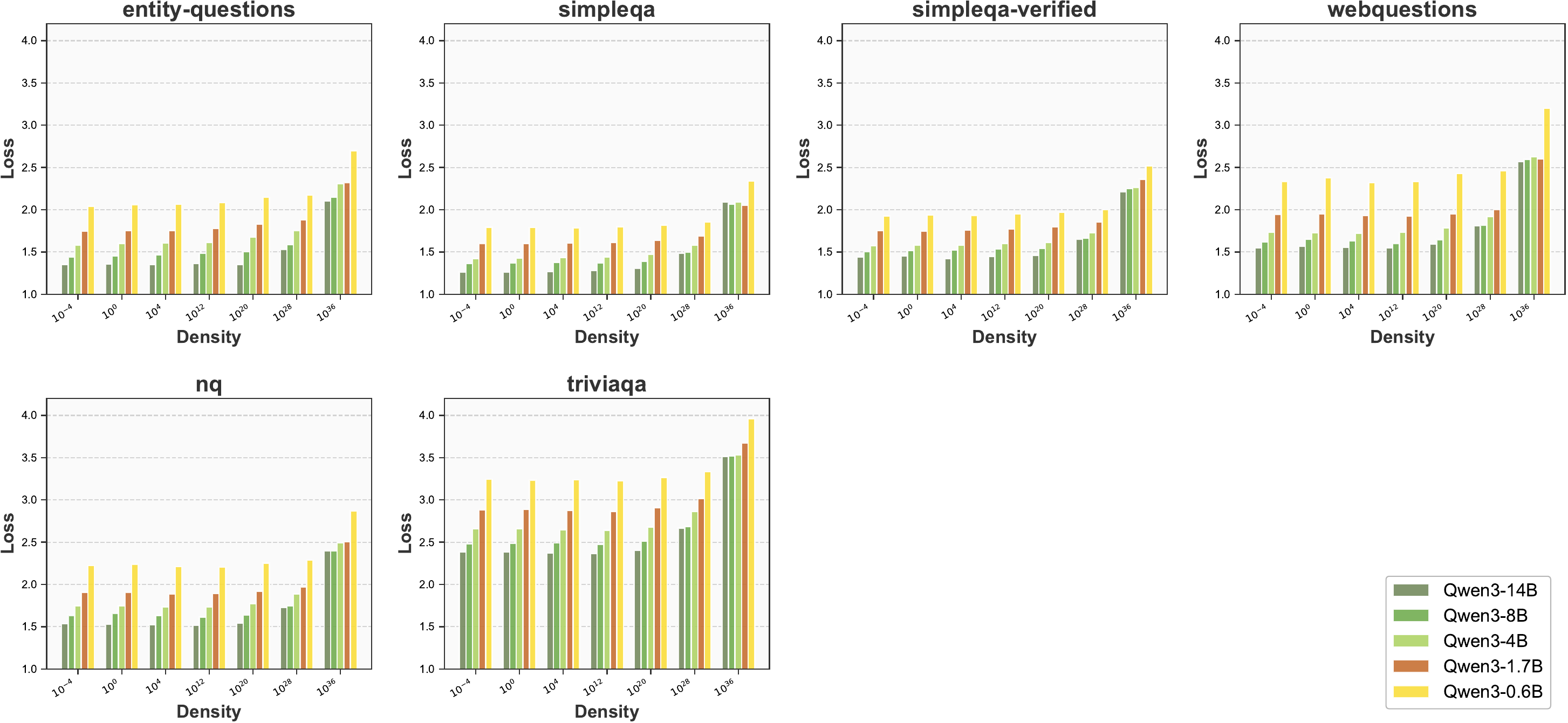}
    \caption{Eval loss of Qwen3-base models of different sizes on Web Questions, Natural Questions, TriviaQA, SimpleQA, SimpleQA-Verified, and EntityQuestions, trained with synthetic data of varying densities.} 
    \label{fig:scaling_model_size_dataset}
\end{figure*}

Below we provide the complete per-dataset loss visualizations for Fig.~\ref{fig:scaling_model_size}, as shown in Fig. \ref{fig:scaling_model_size_dataset}.

\subsection{Complete Case Study}
\label{Complete Case Study}

\begin{figure*}[ht]
    \centering
    \includegraphics[width=1.0\textwidth]{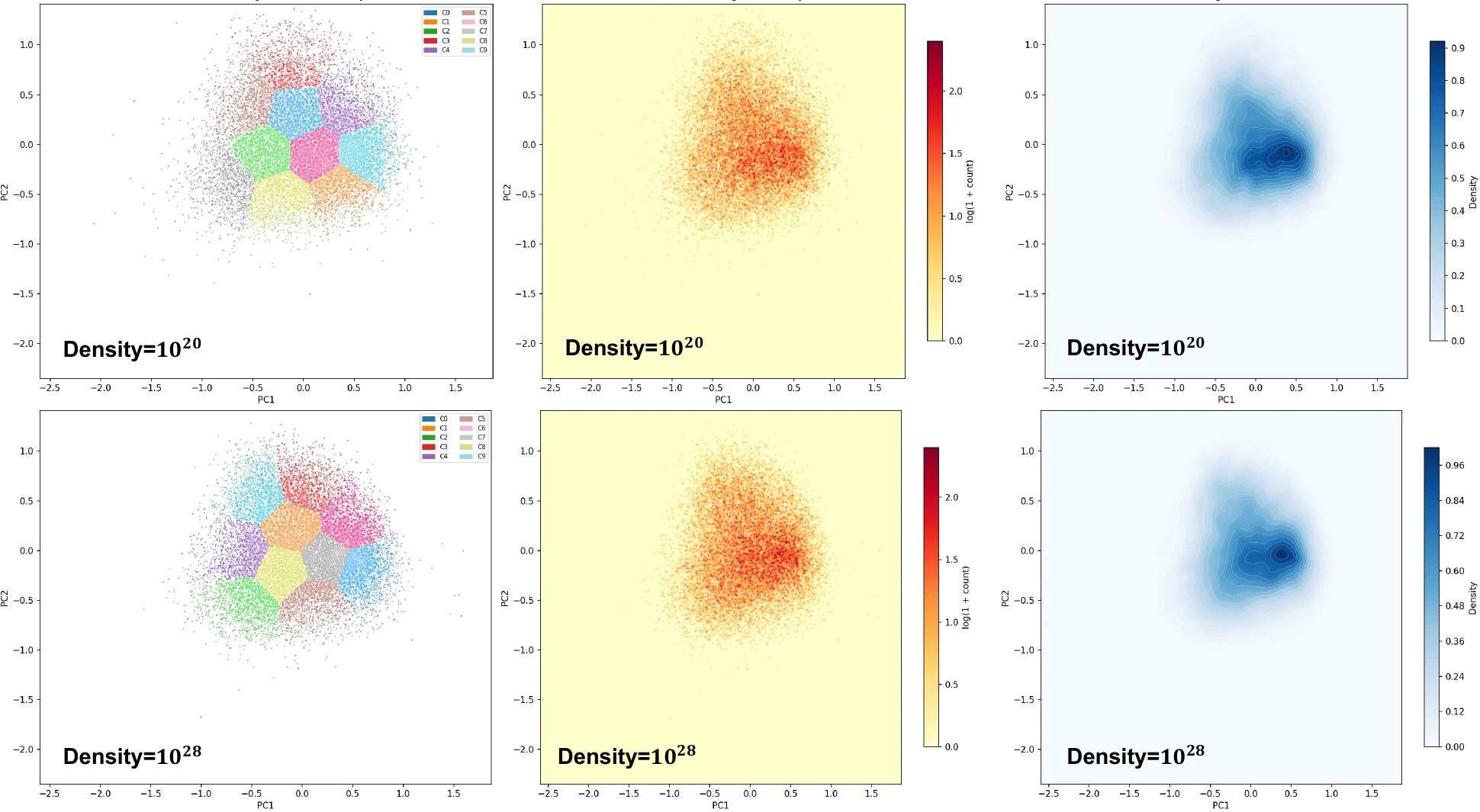}
    \vspace{1mm}
    \includegraphics[width=1.0\textwidth]{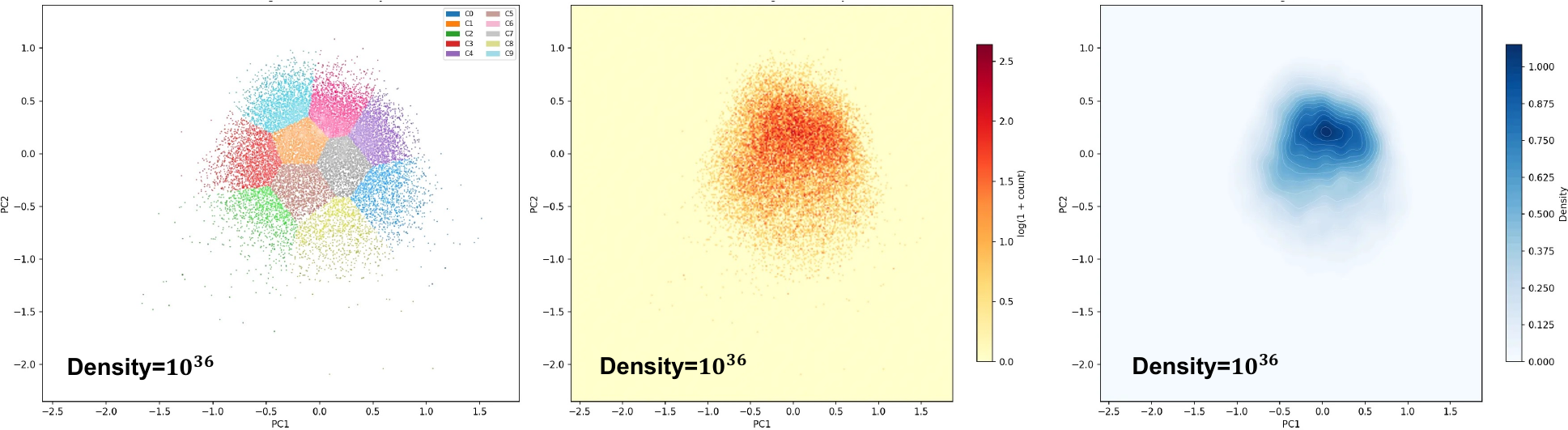}
    \caption{Case study: Visualization of data with different knowledge density distributions.}
    \label{fig:complete_case_study}
\end{figure*}

Below we provide the complete visualizations for Sec.~\ref{Case Study}, covering 7 different knowledge density distributions ranging from $10^{-4}$ to $10^{36}$, as shown in Fig. \ref{fig:complete_case_study}.

\begin{figure*}[t]
    \centering
    \includegraphics[width=1.0\textwidth]{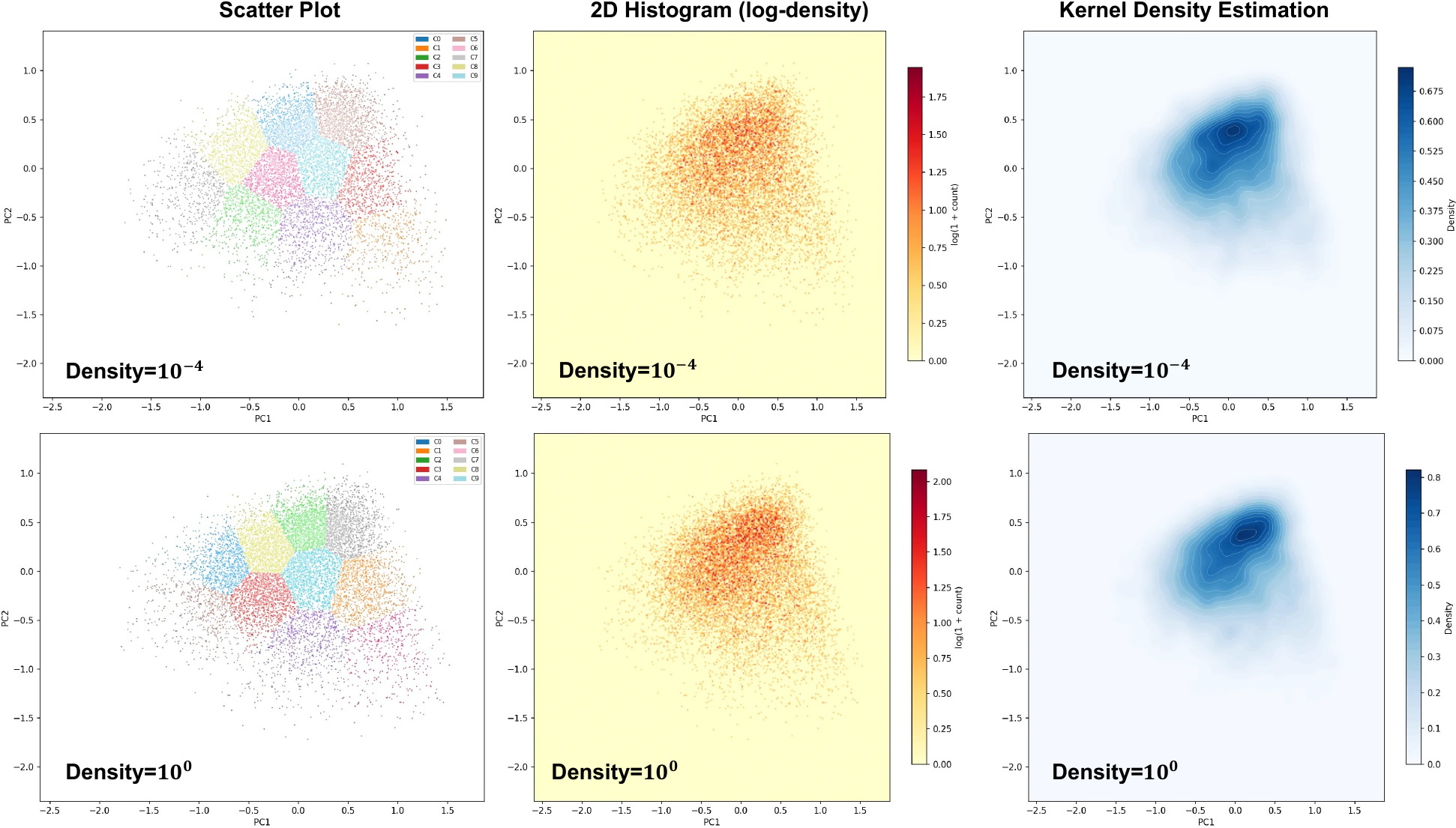}
    \vspace{1mm}
    \includegraphics[width=1.0\textwidth]{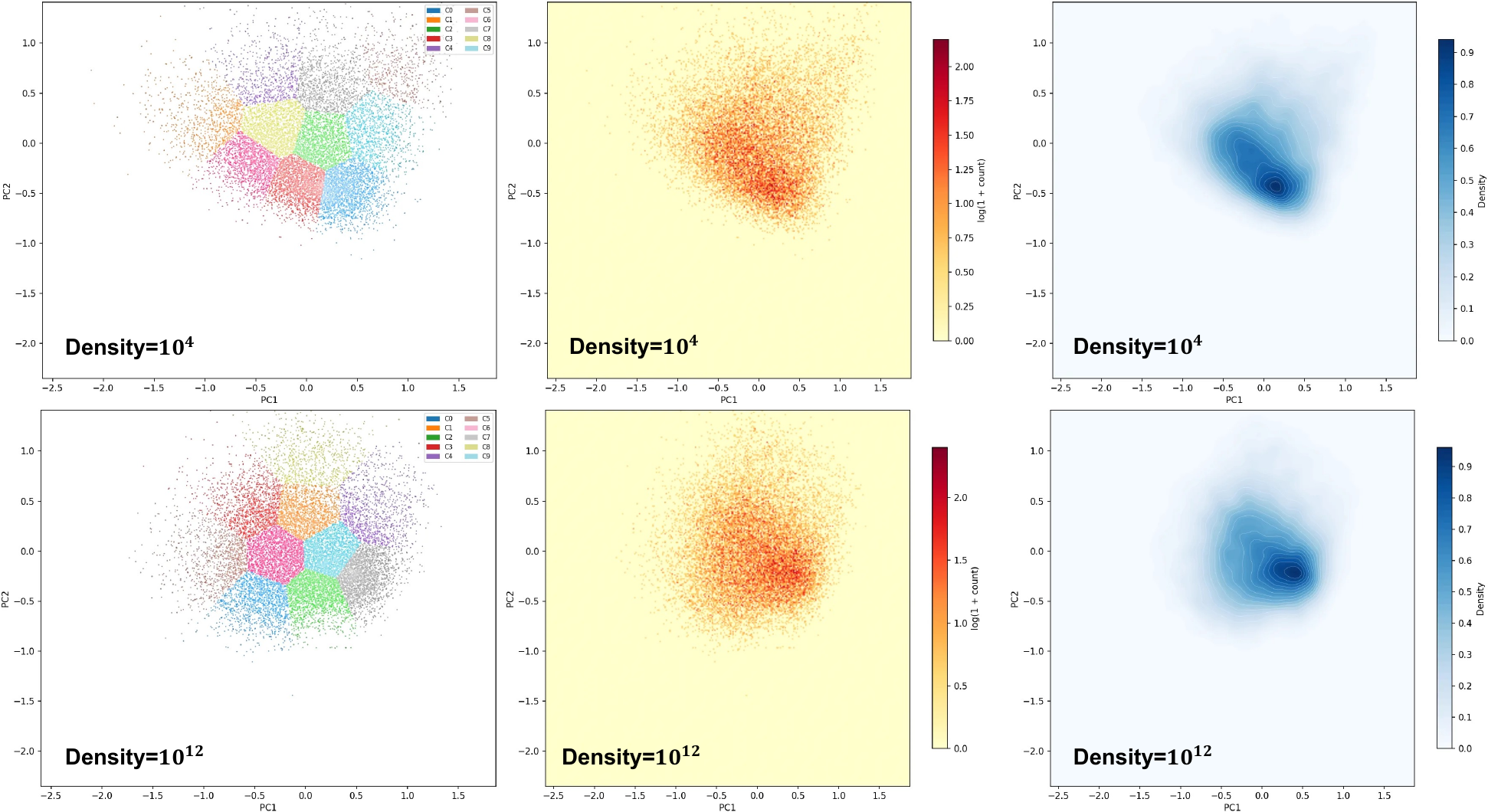}
\end{figure*}

\section{Details of Methods}
\label{Details of Methods}

\subsection{Seed Pool Preparation}
\label{Seed Pool Preparation}
\paragraph{Document Processing.}
We collect raw documents from Wikipedia and apply standard cleaning procedures, including removing tables, reference sections, overly short lines, and separator lines, retaining only the main body text of each article.
\paragraph{Structured Indexing via Knowledge Points.}
Rather than directly chunking documents, we compress each document into a set of concise, knowledge-dense sentences, referred to as \textit{knowledge points}. This avoids common issues with direct chunking such as coreference ambiguity, incomplete information, and uneven length. Each knowledge point is a short, self-contained factual statement that is both amenable to multi-hop extension and suitable for answer verification.
We first extract all named entities from each document, filtering out uninformative types such as dates, quantities, and cardinal numbers. The extracted entities serve as index keys, linking each document to its associated knowledge points and enabling entity-level retrieval across the corpus.
\paragraph{Seed QA Synthesis.}
We adopt two complementary strategies for seed QA synthesis. \textbf{(v1) Document-based synthesis:} We directly prompt an LLM to generate ten factual QA pairs per document based on its content. While this approach produces questions with richer contextual descriptions, it tends to generate more subjective or ambiguous questions. \textbf{(v2) Knowledge-based synthesis:} We sample a random subset of up to 20 knowledge points and prompt an LLM to generate ten QA pairs grounded in those knowledge points. This strategy yields more precise and entity-centric questions, better aligned with the factual and deterministic nature of our evaluation benchmarks. For documents with a large number of knowledge points, we split them into groups of 20 for synthesis. The final seed pool adopts v2 as the primary strategy. In both strategies, each question is required to be at least 30 words, target a specific entity, event, time, or number, and include citations to the involved knowledge point IDs.
\paragraph{Multi-hop Extension.}
To enrich the diversity and complexity of the seed pool, we extend each seed QA via entity-level random walks over the knowledge graph. Starting from the entities in a seed question, we perform random walks of 1--4 hops, collecting the knowledge points associated with the traversed documents. We then merge the top-50 knowledge points (by length) across the traversed documents and prompt an LLM to generate 10 multi-hop questions grounded in these knowledge points, using the original seed question as a reference. Each extended question is required to be at least 50 words and must involve multi-hop reasoning across at least two knowledge points.
\paragraph{Answer Verification.}
Since QA pairs are generated in batches without explicit chain-of-thought reasoning, factual errors are common. We apply an LLM-based verification step to each generated QA pair: given the involved knowledge points, the model is asked to reason through whether the question is valid and whether the answer is correct. Questions that cannot be grounded in the provided knowledge points are discarded; questions with incorrect answers are corrected; and questions deemed unreasonable are also discarded.
This pipeline yields approximately 14M verified seed QA pairs (1.73B tokens), which serve as the input to the KDoS synthesis framework.

\subsection{Knowledge Point Extraction \& Grouping}
\label{Knowledge Point Extraction and Grouping}

Below we present examples of knowledge point lists, knowledge logic chains, and knowledge groups.

\begin{Promptbox}{Case of Knowledge Point Extraction and Grouping}

\textbf{Question:}

A tractor model known for its Perkins 3 cylinder engine and global sales success belongs to a brand for which Agriline provides expert advice and customer service—what is the name of this tractor model?

\textbf{Answer:}

Massey Ferguson 35

\textbf{Knowledge Point List:}

[0] "The Massey Ferguson 35 is a tractor model."

[1] "The Massey Ferguson 35 is equipped with a Perkins 3-cylinder diesel engine."

[2] "The Massey Ferguson 35 is known for its global sales success."

[3] "Agriline is a company that provides expert advice, parts, and customer service for Massey Ferguson tractors."

[4] "Agriline's specialization indicates the brand it supports is Massey Ferguson."

\textbf{Knowledge Logic Chain:}

"graph TD
A[Question] --> B["Tractor model with Perkins 3-cylinder engine"]
B --> C["Perkins 3-cylinder engine is in Massey Ferguson 35"]
A --> D["Tractor with global sales success"]
D --> E["Massey Ferguson 35 is globally successful"]
A --> F["Brand supported by Agriline"]
F --> G["Agriline specializes in Massey Ferguson"]
C --> H["Massey Ferguson 35"]
E --> H
G --> I["Brand: Massey Ferguson"]
I --> H
H --> J["Answer: Massey Ferguson 35"]"

\textbf{Knowledge Group:}

\textbf{Sample 1:}

{

"question": "A tractor model known for its Perkins 3 cylinder engine and global sales success belongs to a brand for which Agriline provides expert advice and customer service—what is the name of this tractor model?",

"answer": "Massey Ferguson 35",

"knowledge\_points": ["The Massey Ferguson 35 is a tractor model.", "The Massey Ferguson 35 is equipped with a Perkins 3-cylinder diesel engine.", "The Massey Ferguson 35 is known for its global sales success.", "Agriline is a company that provides expert advice, parts, and customer service for Massey Ferguson tractors.", "Agriline's specialization indicates the brand it supports is Massey Ferguson."],

"knowledge\_logic": "graph TD    A[Question] --> B[\"Tractor model with Perkins 3-cylinder engine\"]    B --> C[\"Perkins 3-cylinder engine is in Massey Ferguson 35\"]    A --> D[\"Tractor with global sales success\"]    D --> E[\"Massey Ferguson 35 is globally successful\"]    A --> F[\"Brand supported by Agriline\"]    F --> G[\"Agriline specializes in Massey Ferguson\"]    C --> H[\"Massey Ferguson 35\"]    E --> H    G --> I[\"Brand: Massey Ferguson\"]    I --> H    H --> J[\"Answer: Massey Ferguson 35\"]"

}

\textbf{Sample 2:}

{

"question": "The Massey Ferguson 300 series was known for providing a certain type of power and featured a versatile gearbox; which company provides expert advice and customer service on Massey Ferguson parts, including those for this series?",

"answer": "Agriline",

"knowledge\_points": ["The Massey Ferguson 300 series is a line of agricultural tractors known for robust power and a versatile gearbox.", "Massey Ferguson is a brand of agricultural machinery, and its parts often require specialized suppliers for maintenance.", "Agriline is a company specializing in providing parts, expert advice, and customer service for Massey Ferguson machinery.", "The question asks for a company that offers expert advice and customer service on Massey Ferguson parts, specifically referencing the 300 series.", "Agriline is identified as the provider meeting these criteria, based on its known role in the agricultural parts market."],

"knowledge\_logic": "graph TD    A[Massey Ferguson 300 series] --> B{Requires parts/service}    B --> C[Specialized supplier needed]    C --> D[Agriline: expert advice/customer service]    D --> E{Matches question criteria?}    E --> F[Yes: Agriline is answer]"

}

\textbf{Sample 3:}

{

"question": "What organization is mentioned as providing expert advice and customer service on Massey Ferguson parts?",

"answer": "Agriline",

"knowledge\_points": ["Massey Ferguson is a brand of agricultural machinery and equipment.", "Massey Ferguson parts are components used for repairing or maintaining this machinery.", "Organizations may provide services such as expert advice and customer support related to these parts.", "Agriline is a known supplier or service provider specializing in Massey Ferguson parts.", "The question asks for the organization associated with providing expert advice and customer service on these parts."],

"knowledge\_logic": "graph TD    A[Massey Ferguson parts] --> B{Which organization provides<br>expert advice \& customer service?};    B --> C[Agriline];    C --> D((Answer: Agriline));"

}

\end{Promptbox}

\subsection{Algorithm Details}
\label{Algorithm Details}

We detail the rejection sampling procedure from Sec.~\ref{Reject Sampling} in Algorithm~\ref{algorithm:rejection_sampling} below.

\end{document}